\def\year{2022}\relax
\documentclass[letterpaper]{article} 
\usepackage{aaai22}  
\usepackage{times}  
\usepackage{helvet}  
\usepackage{courier}  
\usepackage[hyphens]{url}  
\usepackage{graphicx} 
\usepackage{natbib}  
\usepackage{caption} 
\DeclareCaptionStyle{ruled}{labelfont=normalfont,labelsep=colon,strut=off} 
\usepackage{algorithm}

\usepackage{graphicx}
\usepackage{algpseudocode}

\usepackage{subfig}
\usepackage{wrapfig}
\usepackage[flushleft]{threeparttable}
\usepackage{multirow}

\usepackage{hyperref}
\usepackage[symbol]{footmisc}
\usepackage{amssymb}
\usepackage{amsthm}
\usepackage{amsmath}   
{
      \theoremstyle{plain}

      \newtheorem{defn}{Definition}

      \newtheorem{remark}{Remark}
      \newtheorem{prop}{Proposition}
}

\usepackage{mathtools}

\newcommand{\ceil}[1]{\left\lceil #1 \right\rceil}

\makeatletter
\DeclareRobustCommand*\cal{\@fontswitch\relax\mathcal}
\makeatother

\usepackage{newfloat}
\usepackage{listings}
\floatstyle{ruled}
\newfloat{listing}{tb}{lst}{}
\floatname{listing}{Listing}

\title{\texttt{AET-SGD:} Asynchronous Event-triggered Stochastic Gradient Descent}

\author{
     Nhuong Nguyen, Song Han}
\affiliations{
Department of Computer Science and Engineering, University of Connecticut \\
\{nhuong.nguyen, song.han\}@uconn.edu}

\usepackage{bibentry}

\begin{document}

\maketitle

\begin{abstract}
Communication cost is the main bottleneck for the design of effective distributed learning algorithms. Recently, event-triggered techniques have been proposed to reduce the exchanged information among compute nodes and thus alleviate the communication cost. However, most existing event-triggered approaches only consider heuristic event-triggered thresholds. They also ignore the impact of computation and network delay, which play an important role on the training performance. In this paper, we propose an Asynchronous Event-triggered Stochastic Gradient Descent (SGD) framework, called \texttt{AET-SGD}, to i) reduce the communication cost among the compute nodes, and ii) mitigate the impact of the delay.
Compared with baseline event-triggered methods, \texttt{AET-SGD} employs a linear increasing sample size event-triggered threshold, and can significantly reduce the communication cost while keeping good convergence performance. 
We implement \texttt{AET-SGD} and evaluate its performance on multiple representative data sets, including MNIST, FashionMNIST, KMNIST and CIFAR$10$. The experimental results validate the correctness of the design and show a significant communication cost reduction from $44$x to $120$x, compared to the state of the art.  
Our results also show that \texttt{AET-SGD} can resist large delay from the straggler nodes while obtaining a decent performance and a desired speedup ratio.
\end{abstract}

\section{Introduction}


The optimization problem for training many machine learning (ML) models using a training set $\{\xi_i\}_{i=1}^m$ of $m$ samples can be formulated as a general finite-sum minimization problem as follows
\begin{equation}\label{eq:finite_sum_main}
\min_{w \in \mathbb{R}^d} \left\{ F(w) = \frac{1}{m}
\sum_{i=1}^m f(w; \xi_i) \right\}.
\end{equation}
The objective is to minimize a loss function $f$ with respect to model parameters $w \in \mathbb{R}^d$, where $d$ represents the model dimension. This problem is known as empirical risk minimization and it covers a wide range of convex and non-convex problems 
in 
the ML domain, including but not limited to logistic regression, multi-kernel learning, conditional random fields and neural networks~\citep{roux2012stochastic,bottou2018optimization,nguyen2018sgd}. In this paper, we aim to solve the following more general stochastic optimization problem with respect to some distribution $\mathcal{D}$, which is a conventional way to describe the general empirical and expected risk minimization problem:
\begin{align}
\min_{w \in \mathbb{R}^d} \left\{ F(w) = \mathbb{E}_{\xi \sim \mathcal{D}} [ f(w;\xi) ] \right\},  \label{eq:eqObj}  
\end{align}
where $F$ has a Lipschitz continuous gradient and $f$ is bounded from below for every $\xi$.

Different from most existing solutions which are based on a centralized setting~\citep{stich2018local,yang2019,konevcny2016federated}, we aim to solve (\ref{eq:eqObj}) in a distributed fashion where compute nodes (or clients
) have their own local data sets and the minimization problem is solved over the collection of all local data sets. A widely adopted approach to solve this problem is to repeatedly use the following SGD recursion~\citep{robbins1951stochastic}
\begin{equation}
 w_{t+1} = w_t - \eta_t  \nabla f(w_t;\xi),\label{eqwSGD}
 \end{equation}
where $w_t$ represents the model after the $t$-th iteration and $\eta_t$ is the step size at iteration $t$; $w_t$ is used in computing the gradient of $f(w_t;\xi)$, where $\xi$ is a data sample randomly selected from the data set $\{\xi_i\}_{i=1}^m$ which comprises the union of all local data sets.

The iterative update~(\ref{eqwSGD}) allows us to paralelize the computation by using the client--server 
architecture. 
Specifically, this approach allows each client to perform a certain number of local SGD recursions for $\xi$ that belongs to the client's local data set. The sum of gradient updates ${\cal U}$\footnote{The sum of gradient ${\cal U}$ represents the accumulative gradient that clients compute within a local SGD round.} as a result of the SGD recursion (\ref{eqwSGD}) is sent to a centralized server which aggregates all received updates ${\cal U}$ from clients and maintains a global model. The server regularly broadcasts its most recent global model so that clients can use the new model in their local SGD computations. This allows each client to use what has been learnt from the local data sets at the other clients, leading to good accuracy of the final global model.
Based on the way that the clients update their local SGD computations, we can classify the above client--server distributed learning approach into synchronous SGD (SSGD)~\citep{jianmin,mcmahan,stich2018local,bonawitz2019towards}, where the compute nodes need synchronous points for every communication round and asynchronous SGD (ASGD)~\citep{zinkevich2009slow,lian2015asynchronous,zheng2017asynchronous,meng2017asynchronous,stich2018local,shi2019distributed}, where compute nodes can asynchronously update their computed information without any synchronization.
The line of work for distributed learning
confirms that SSGD can obtain a good convergence rate while its training time is significantly affected by the synchronous points, i.e., each client needs to wait for the slowest client. On the other hand, ASGD can fully utilize the computational power of all devices, but it is hard to guarantee the convergence rate~\citep{jianmin,stich2018local,van2020hogwild,zheng2017asynchronous,meng2017asynchronous}.

\begin{figure}[ht!]
\vspace{-.25cm}
  \centering
  \subfloat[Client--Server architecture.]{\includegraphics[width=0.25\textwidth]{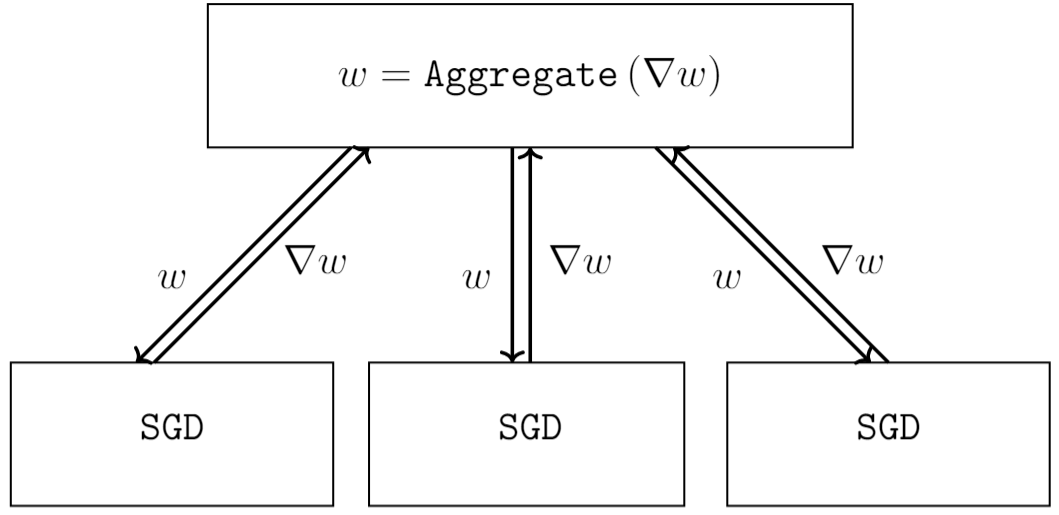}\label{fig:central_server}}
  \hfill
  \subfloat[P2P architecture.]{\includegraphics[width=0.199\textwidth]{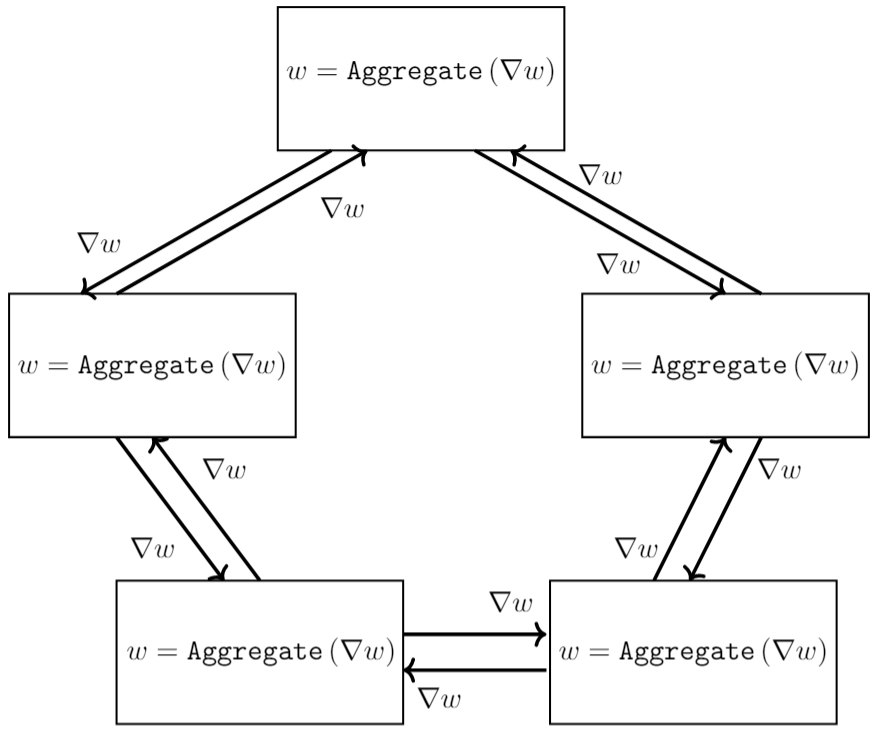}\label{fig:ring_topology}}
  \caption{Comparison of (\ref{fig:central_server}) client--server architecture and (\ref{fig:ring_topology}) peer--to--peer (P2P) architecture. In the client-server architecture, compute nodes use local SGD method and exchange the gradient information with the central server; in the peer--to--peer architecture, the gradient information is only exchanged with the compute nodes' neighbors.
  }
  \label{fig:network_topology2}
\end{figure}

While SSGD and ASGD work mainly with the client--server undirected network topology as shown in Figure~\ref{fig:central_server},
in recent years, event-triggered SGD frameworks~\citep{hsieh2017gaia,luping,george2019distributed} are proposed to adapt to any kind of network topology,
and can work well under asynchronous settings. Event-triggered SGD allows the clients to independently compute their local SGD recursions and broadcast their SGD computations to only their neighbors (e.g., as a ring network topology) when a predefined event-triggered threshold is satisfied,
as shown in Figure~\ref{fig:ring_topology}.
However, the threshold definition is usually based on heuristic choice, so its performance can be deteriorated by a badly selected event threshold. Furthermore, the baseline event-triggered SGD does not analyze the network delay, which is an important factor affecting the training performance. 
This makes 
the existing event-triggered SGD
frameworks less practical when deployed in real-life scenarios. To address these weakness,
in this paper, we propose a new asynchronous event-triggered SGD framework, called \texttt{AET-SGD}, which inherits the nice features of event-triggered SGD with local SGD, to not only be  able to work under all kinds of network topology but 
reduce the communication cost among all clients. Besides, \texttt{AET-SGD} is carefully designed to tolerate unexpected network delays.

Our approach is based on the Hogwild!~~\citep{Hogwild, van2020hogwild}
recursion 
\begin{equation}
 w_{t+1} = w_t - \eta_t  \nabla f(\hat{w}_t;\xi_t),\label{eqwM2a}
 \end{equation}
 where $\hat{w}_t$ represents the vector used in computing the gradient $\nabla f(\hat{w}_t;\xi_t)$, $\eta_t$ is the step size at iteration $t$,
in which
 vector entries have been read (one by one)  from  an aggregate of a mix of  previous updates that led to $w_{j}$, $j\leq t$.
 

Recursion (\ref{eqwM2a}) models asynchronous SGD, and defines the amount of asynchronous behavior as the following function $\tau(t)$:

\begin{defn} \citep{van2020hogwild} \label{def:delay}
We say that the sequence $\{w_t\}$  is consistent with a delay function $\tau$  
if, for all $t$, vector $\hat{w}_t$ includes the aggregate 
of the updates up to and including those made during the $(t-\tau(t))$-th iteration\footnote{ (\ref{eqwM2a}) defines the $(t+1)$-th iteration, where $\eta_t\nabla f(\hat{w}_t;\xi_t)$ represents the $(t+1)$-th update.}, i.e., $\hat{w}_t = w_0 - \sum_{j\in {\cal U}} \eta_j \nabla f(\hat{w}_j;\xi_j)$ for some ${\cal U}$ with $\{0,1,\ldots, t-\tau(t)-1\}\subseteq {\cal U}$.
\end{defn}

The asynchronous SGD framework based on Hogwild! can resist 
the network delay
caused by the network communication infrastructure, $\tau(t)$ $\approx \sqrt{t/\ln t}$, where $t$ is the current SGD iteration~\citep{nguyen2018sgd}.
Hence, we can judiciously exploit this delay property. Specifically, we can {\em increase the number of SGD iterations performed locally at each compute node  from round to round} and {\em adapt this setting to the specific network topology}. This design {reduces the amount of network communication} compared to the straightforward usage of recurrence (\ref{eqwM2a}) where each compute node performs a constant number of SGD iterations within each round. 

To the best of our knowledge, there does not exist any distributed SGD framework in the literature for non-convex optimization problems that utilize the \textit{linearly increasing sample sequences} and can work well under \textit{peer-to-peer} network topology.
This setting will help significantly save the resources used for communication. Moreover, our proposed framework 
is motivated by
the theoretical analysis from Hogwild! framework~\citep{nguyen2018sgd,van2020hogwild}, which guarantees that \texttt{AET-SGD} can converge to the stationary solutions for non-convex problems.





The key contributions of our work are summarized below:
\begin{enumerate}
    \item We propose a new asynchronous event-triggered SGD framework, called \texttt{AET-SGD}, which can significantly reduce the communication cost among clients by using {\em linearly increasing local SGD recursions} as the event-triggered threshold, compared to other constant local SGD recursions or heuristic event-triggered threshold. Moreover, \texttt{AET-SGD} can work well under a fully distributed setting with a convergence guarantee (to the stationary solutions) for non-convex objective functions.
    
    \item We introduce the definition and implementation of delay property $\tau$, the maximum asynchronous behaviours that the event-triggered SGD framework can tolerate without affecting the final convergence performance. From the theoretical perspective~\citep{nguyen2018sgd,nguyen2018new}, each client can work well with the delay $\tau(t) \approx \sqrt{t / \ln{t}} $. Here, $t$ is the current iteration and our clients use the diminishing step size scheme. {\em To the best of our knowledge, we are the first to show the design and analysis of peer-to-peer learning system from the perspective of Hogwild! algorithm, which not only conserves the convergence, but also mitigates the effect of delay.}
    
    \item We comprehensively evaluate the convergence performance of \texttt{AET-SGD} for non-convex problems, and confirm the robustness of our framework in terms of good test accuracy. 
    
    
    
     \item{{We also evaluate the performance of \texttt{AET-SGD} 
    at
     different levels of asynchrony. The experimental results show that \texttt{AET-SGD} can alleviate the effect of stragglers, i.e., nodes that are significantly slower than average and have an unstable network condition.}}
\end{enumerate}



\section{Related work}
\label{sec:related_work}





From the perspective of synchronous points, the distributed learning framework (based on SGD) can be divided into two main lines of research: synchronous SGD (SSGD) and asynchronous SGD (ASGD).

\vspace{0.025in}
\noindent \textbf{SSGD.} Local SGD~\citep{stich2018local} and Federated Learning (FL)~\citep{jianmin,mcmahan,bonawitz2019towards} are distributed machine learning approaches which enable training on a large corpus of decentralized data located on devices like mobile phones or IoT devices. While local SGD does not care about the data privacy, FL assumes that there is no data shared between any client to prevent data leakage. Original FL requires synchrony between the server and clients.  It requires that each client sends a full model back to the server in each round and each client needs to wait for the next computation round. For large and complicated models, this becomes a main bottleneck  due to the asymmetric property of network connections and the different computation power of devices~\citep{yang2019,konevcny2016federated}. Distributed mini-batch SGD~\citep{jianmin,you2018imagenet} is a special case of SSGD framework, where each compute node exchanges its gradient or local model for every single (mini-batch) SGD recursion. However, the problem of \textit{generalization} needs to be considered when using large mini-batch sizes
~\citep{you2017scaling,lin2018don,yin2018gradient}.

\vspace{0.025in}
\noindent \textbf{ASGD.} Asynchronous SGD ~\citep{zinkevich2009slow,lian2015asynchronous,zheng2017asynchronous,meng2017asynchronous,stich2018local,shi2019distributed} can be used under the distributed learning setting. Hogwild!, one of the most famous ASGD algorithms, was introduced in ~\citep{Hogwild} and various variants with a fixed or diminishing step size sequences were introduced in \citep{dean2012large,ManiaPanPapailiopoulosEtAl2015,DeSaZhangOlukotunEtAl2015,Leblond2018,nguyen2018sgd}. 
%
Typically, asynchronous training converges faster than synchronous training 
due to parallelism. This is because in a synchronized solution compute nodes have to wait for the slower ones to communicate their updates after a new global model can be downloaded by everyone. This causes high idle waiting times at individual compute nodes.
By contrast, asynchronous training allows compute nodes to continue executing SGD recursions based on stale global models.


In \citep{cong2019,yang2019} asynchronous training combined  with federated optimization is proposed. Specifically, the server and clients conduct updates asynchronously: the server immediately updates the global model whenever it receives a local model from clients. Therefore, the communication between the server and clients is non-blocking and more effective.
\citep{cong2019} also provides a convergence analysis under the asynchronous setting. 
The work from \citep{tianli2019hetero} introduces FedProx which is a modification of FedAvg (i.e., original FL algorithm of \citep{mcmahan}). In FedProx, the clients solve a proximal minimization problem rather than traditional minimization as in FedAvg. 

Another work~\citep{van2020hogwild} introduces asynchronous SGD with increasing sample sizes to 
reduce the communication cost. This work analyses and demonstrates the promise of 
linearly increasing sample sequence technique
with the convergence rate guarantee for both iid and heterogeneous data sets. Another work~\citep{haddadpour2019local} also utilizes the benefit of increasing sample size over the communication rounds, but it only focuses on the {synchronous} setting. 

\vspace{0.025in}
\noindent \textbf{Event-triggered SGD} is a special case of ASGD, where each client independently computes and broadcasts the SGD computation to its neighbors. The event-triggered SGD framework can be combined with synchronization barrier to synchronize the SGD updates among all clients. Hence, we classify the event-triggered SGD framework into \textit{synchronous} event-triggered SGD and \textit{asynchronous} event-triggered SGD.
In this paper, we focus on \textit{asynchronous} event-triggered SGD framework. The event-triggered threshold also plays an important role in this algorithm since it decides the amount of communication cost among clients and the convergence performance. While the work~\citep{george2019distributed} measures the difference between the \textit{current local model} and the \textit{previous one} as a threshold to broadcast the current model, other studies~\citep{hsieh2017gaia,luping} use the relevant gradient metric from the clients to decide the period when the SGD computations can be sent to other neighbors. However, these approaches are proposed and analyzed without considering the network delay,
which restricts their practicability in real-life applications.

In this paper, we extend the event-triggered framework
by incorporating in the linearly increasing sample sequence
and a corresponding delay measurement. 
In other words, our event-triggered threshold is the period when the compute nodes finish their local SGD iterations with respect to the current communication round, instead of the heuristic event-triggered threshold as usual.
Compared to the existing solutions using the linearly increasing sample size technique~\citep{haddadpour2019local,van2020hogwild}, \texttt{AET-SGD} focuses on the peer-to-peer network architecture 
with network delay constraint and is able to work well in {\em heterogeneous environment with straggler nodes}.


\section{Preliminaries and Motivation}
\label{sec:preliminary_motivation}



\subsection{Event-triggered SGD}


As described above, event-triggered SGD is a type of ASGD, which can run for any kinds of network topology. Specifically, when an event-triggered threshold is satisfied, the clients will broadcast the local SGD computations to their neighbors. The event-triggered threshold decides the convergence performance and the communication cost among the clients.
The event-triggered threshold is usually measured by the norm of distance between the current model $w_t$ and the previous model $w_{t-1}$
, where $t$ is the current SGD iteration,
and the number of model parameters, denoted as $N_{p}$~\cite{george2019distributed}.
{To understand how the event-triggered threshold affects the convergence performance, we conduct an experiment on MNIST data set with $5$ compute nodes, forming the ring network topology as shown in Figure~\ref{fig:ring_topology}.
Also, each compute node will run for a total of $10,000$ iterations.}
The communication event or communication round is triggered when the norm distance exceeds the predefined threshold. Here, $\eta_t$ is the current step size at SGD iteration $t$.
As can be observed from Table~\ref{tbl:event_triggered_threshold}, the test accuracy of event-triggered SGD 
increases along with the decrease of the threshold,
which indicates that the event-triggered threshold is a key factor of
the training performance and needs to be carefully selected to achieve good tradeoff between the test accuracy and the communication overhead. {We also consider a simple linear increasing sample sequence $s_i = 10 \cdot i$ (as threshold\_6), where $i$ is the current communication round, which only takes $46$ communication rounds in total while achieving a good test accuracy.}
Moreover, the event-triggered threshold selection in the literature is usually heuristic~\citep{hsieh2017gaia,luping,george2019distributed}, so it is hard to measure the communication cost and provide the theoretical analysis. To address this issue, in this work we exploit the linearly increasing sample sequences technique from round to round to reduce the communication cost and obtain a decent test accuracy.
\begin{table}[t]
\centering
\scalebox{0.8}{
\begin{tabular}{|c|c|c|c|}
\hline
Name         & Threshold & \# of round & Test accuracy \\ 
\hline
\hline
threshold\_1 &   $\leq \eta_t \cdot N_{p}$
&  $214$            &   $0.8985$                
\\ \hline
threshold\_2 &  $\leq 0.8 \cdot \eta_t  \cdot N_{p}$     &   $345$           &   $0.9162$                \\ \hline
threshold\_3 & $\leq 0.6 \cdot \eta_t \cdot N_{p}$      &    $428$          &    $0.9303$               \\ \hline
threshold\_4 &  $\leq  0.4 \cdot \eta_t \cdot N_{p}$     &    $687$         &     $0.9576$              \\ \hline
threshold\_5 &  $\leq 0.2 \cdot \eta_t \cdot N_{p}$     &    $1085$          &        $0.9673$          
\\ \hline \hline

threshold\_6 &  \texttt{Increasing}   &    $46$
&        $0.9702$
\\
\hline 
\end{tabular}
}
\vspace{-0.075in}
\caption{Communication cost and test accuracy of event-triggered SGD with different thresholds.}
\label{tbl:event_triggered_threshold}

\end{table}

\subsection{Hogwild! Algorithm}
    

Hogwild! algorithm~\citep{Hogwild,nguyen2018sgd} is a well-known asynchronous distributed learning method. Compared to the event-triggered SGD, Hogwild! mainly works with the client--server network topology
as shown in Algorithm~\ref{alg:HogWildAlgorithm}.
Moreover, Hogwild! is provable to be able to tolerate the network delay $\tau(t) \approx \sqrt{t/\ln{t}}$, where $t$ is the current SGD iteration~\citep{nguyen2018sgd}. It means that Hogwild! can resist a larger network delay 
when the number of SGD iteration $t$ increases. In other words, the gradient updates from each compute nodes can endure more delay, leading to the linearly increasing sample size from round to round without deteriorating the convergence performance\footnote{See Theorem $5$ of~\citep{nguyen2018sgd} for details of mathematical explanation.}. 

\begin{algorithm}[t]
\caption{Hogwild! general recursion}
\label{alg:HogWildAlgorithm}
\begin{algorithmic}[1]
   \State {\bf Input:} $w_{0} \in \mathbb{R}^d$
   \For{$t=0,1,2,\dotsc$ {\bf in parallel}} 
    
  \State read each position of shared memory $w$
  denoted by $\hat w_t$  {\bf (each position read is atomic)}
  \State draw a random sample $\xi_t$ and a random ``filter'' $S^{\xi_t}_{u_t}$
  \For{positions $h$ where $S^{\xi_t}_{u_t}$ has a 1 on its diagonal}
   \State compute $g_h$ as the gradient $\nabla f(\hat{w}_t;\xi_t)$ at position $h$
   \State add $\eta_t d_{\xi_t} g_h$ to the entry at position $h$ of $w$ in shared memory {\bf (each position update is atomic)}
   \EndFor
   \EndFor
\end{algorithmic}
\end{algorithm}

In this work, we exploit the property of delay function $\tau(t)$ and the linearly increasing 
sample size sequences with the convergence guarantee.


\subsection{System model}
\label{sec:system_model}

\subsubsection{Network Setting}
\label{subsec:network_setting}

In this work, we employ the undirected network topology.
We assume that messages/packets are never lost; if dropped, they will be resent but can arrive out of order, i.e. the messages are guaranteed to deliver with potential network delay.
In practice, we can detect the list of compute node's neighbors based on their locations and locally form the communication group~\citep{sheng2020learning}. In the scope of this work, we 
assume
that the communication group of each compute node, i.e, the list of neighbors for each compute node is available.

\subsubsection{System delay}
\label{subsec:network_delay} 

In our setting, we simulate the compute node that might have both computation and communication delays. In other words, we randomly generate the delay in the compute node to simulate the asynchronous round. The detailed setting can be found in the experimental section.

\section{Design and Analysis of \texttt{AET-SGD}}
\label{sec:framework}
We now introduce \texttt{AET-SGD}, a new variant of asynchronous event-triggered SGD framework, which uses the linearly increasing sample sequences with diminishing step size.
\subsection{Linearly increasing sample sequences}
\label{sec:sample_sequence}
From the work~\citep{nguyen2018sgd,nguyen2018new,van2020hogwild}, we can suitably choose a delay function $\tau(t) \approx \sqrt{t / \ln{t}}$ (See Definition~\ref{def:delay}), so that we have the sample size sequence\footnote{See Section~$4.1$ of~\citep{van2020hogwild} for obtaining an increasing sample size sequence given a delay function $\tau$.}
as follows
\begin{equation} \label{eq:sample_sequence}
    s_i  = \Theta \left({i} / {\ln i} \right )
\end{equation}
For a fixed number of gradient computations $K$, the number $T$ of communication rounds satisfies $K=\sum_{j=0}^T s_j$. 
The $\ln{i}$ component in~(\ref{eq:sample_sequence}) is usually negligible, compared to the value of $i$.
This makes $T$ proportional to $\sqrt{K}$, rather than proportional to $K$ for a constant sample size sequence. 
From theory and simulation~\citep{van2020hogwild}, in order to bootstrap convergence, it is best  to start the training process with larger step sizes and start with  rounds of small size (measured in the number of local SGD recursions). After that, these rounds should start increasing in  size and should start using smaller and smaller step sizes for best performance (in terms of minimizing communication cost).
In practice, for non-convex problems, we can choose the sample size sequence $s_i = \mathcal{O}(i)$, where $i$ is the current index of the communication round and the corresponding diminishing step size sequence follows ${\eta_i} \sim \mathcal{O}(1 / \sqrt{t})$, where $t$ is the number of iterations until communication round $i$. Note that we can keep a constant step size ${\eta_i}$ within the local sample size sequence $s_{i,c} = \frac{s_i}{n}$, where $n$ is the number of compute nodes joining the training process.

\subsection{Asynchronous Event-triggered SGD}
\begin{algorithm}[ht!]
\scriptsize
\caption{Initial Setup of \texttt{AET-SGD}~\citep{van2020hogwild}} 
\label{alg:setup}
\begin{algorithmic}[1]
\Procedure{Setup}{$n$}
    \State {\bf Initialize} global model $\hat{w}_0=\hat{w}_{c,0,0}$ for all compute nodes $c\in \{1,\ldots, n\}$, diminishing round step size sequence $\{\bar{\eta}_{i}\}_{i\geq 0}$, and increasing sample size sequence $\{s_{i}\}_{i\geq 0}$
    \For{$i \geq 0$}
        \For{$t \in \{0,\ldots, s_i-1\}$}
            \State Assign $a(i,t)=c$ with probability $p_c$
        \EndFor
        \For{$c\in \{1,\ldots n\}$}
            \State $s_{i,c}= |\{ t \ : \ a(i,t)=c\}|$
        \EndFor
    \EndFor
    \State \Comment{$\{s_{i,c}\}_{i\geq 0}$ represents the sample size sequence for compute node $c$; notice that $\mathbb{E}[s_{i,c}]=p_c s_i$}
    \State {\bf Initialize} permissible delay function $\tau(\cdot)$ with $t-\tau(t)$ increasing in $t$
\EndProcedure
\end{algorithmic}
\end{algorithm}


\begin{algorithm}[ht!]
\scriptsize
\caption{\texttt{AET-SGD} Framework}
\label{alg:client_adaptive}
\begin{algorithmic}[1]
\Procedure{Receive}{$message$} \Comment{Parallel with \textsc{MainComputeNode}}
\If{$message(v, \mathcal{U}, k)$ comes from other's neighbors}



\State Update $w_{c} = w_{c} - \bar{\eta}_k \cdot \mathcal{U}$


\State Update $\mathcal{H}_{v} = \mathcal{H}_{v} + 1$ \Comment{update the remote round of neighbor $v$}

\EndIf
\EndProcedure

\State
\Procedure{Synchronize}{$\mathcal{H}, d, i$} 


\State{$d_c = 0$} 
\For{ $e \in \mathcal{H}$ }
    \State{ $d_c = \max \left \{ {d_c, (i - \mathcal{H}_e)} \right \}$ }
\EndFor

\If{$d_c > d$}
    \State{ Return \textsc{True}} 
    
\EndIf

\State{ Return \textsc{False}}

\EndProcedure

\State
\Procedure{MainComputeNode}{$\mathcal{D}_c$, $d$}

\State Initialize model $w_c = \hat{w}_{c,0,0}$ \Comment{Initial model from \Call{Setup}{}}

\State Initialize the map $\mathcal{H}$ and set $\mathcal{H}_{j \neq c } = 0$ if $c, j$ are neighbors.


\State  Initialize $\mathcal{U}_{i} = 0$ and the number of SGD sequence $s_{0,c}$

\For{round $i=0$ to $T$}
    \State $h=0, \mathcal{U}_{i} = 0$ 
    
    
    \While{$h < s_{i,c}$}
    
        \While{\textsc{Synchronize}($\mathcal{H}$, $d$, $i$) == \textsc{True}}  \Comment{The delay checkpoint}
        \State{Only keep tracking the value of $\mathcal{H}$}
        \EndWhile
        
        \State Sample uniformly at random $\xi$ from $\mathcal{D}_c$
        
        \State $g = \nabla f(w_c, \xi)$
        
        \State $w_c = w_c - {\bar{\eta}}_{i} \cdot g$  
        
        
        \State $\mathcal{U}_{i} = \mathcal{U}_{i} + g$ 
        
        \State $h = h + 1$
    \EndWhile
    
    
     \State Detect neighbors of $c$ 
     
    
    \State Send a tuple $(c, \mathcal{U}_{i}, i)$ to the other $c'$s neighbors
    
    
    \State Update a new local SGD sequence $s_{i,c}$
     
\EndFor
   
\EndProcedure

\end{algorithmic} 
\end{algorithm} 
The proposed \texttt{AET-SGD} framework is summarized in Algorithm~\ref{alg:setup} and Algorithm~\ref{alg:client_adaptive}. Algorithm~\ref{alg:setup} presents the set up phase, where we initialize the sample size sequence and the corresponding step size.
Algorithm~\ref{alg:client_adaptive} provides the detailed algorithm which runs on the compute nodes.
The algorithms have the following main procedures.

\vspace{0.025in}
\noindent \textbf{\textsc{Setup} procedure.} We remark that the sample sequence $\{s_{i,c}\}$ is initialized by a coin flipping procedure 
in Algorithm~\ref{alg:setup}. Since the sequence $\{s_i\}$ is increasing, we may approximate $s_{i,c}\approx \mathbb{E}[s_{i,c}]=p_c s_i$ (because of the law of large numbers). Here, $s_i$ is the global number of SGD iterations among all compute nodes at round $i$ and $s_{i,c}$ is the local number of SGD iterations at node $c$ with probability $p_c$. Our setting allows the training process to work under heterogeneous setting, where the compute nodes might have unequal computation capacity and unequal number of local SGD recursions. For simplicity, we can uniformly set the probability $p_c = \frac{1}{n}$, where $n$ is the number of compute nodes. It indicates that each compute node has the same number of local SGD recursions.

\vspace{0.025in}
\noindent \textbf{\textsc{MainComputeNode} procedure.} This function works as the core function with $3$ purposes, including \textit{running local SGD recursions, checking delay and exchanging gradient information} in Algorithm~\ref{alg:client_adaptive}. The compute nodes will run the internal local SGD recursions from round to round. Moreover, the \textsc{Receive} procedure will run parallel with the main \textsc{MainComputeNode} procedure. We also introduce the \textsc{Synchronize} procedure, which simply checks the current communication round among compute nodes based on the communication history $\mathcal{H}$ and keeps the training process consistent to the delay function $\tau$.

Specifically, each compute node will run the local SGD recursions for a batch of local data. These recursions together represent a local round and at the end of the local round, the sum of local model update $\mathcal{U}$
is transmitted to its neighbors. The neighbors in turn add the received sum of local update to its current model $w$ and update the communication history.
As long as the delay checkpoint is satisfied, the compute node will continue its computation without any waiting time. Otherwise, the waiting loop will be activated until the delay checkpoint \textsc{Synchronize} is satisfied. 

\vspace{0.025in}
\noindent \textbf{\textsc{Receive} procedure.} As soon as the compute nodes receive the sum of gradient from their neighbors, this sum of gradient will be updated into their local model ${w}$ with the step size of the current round. 
The current communication round $k$ of the neighbors will also be recorded to the history as show in line $4$ of Algorithm~\ref{alg:client_adaptive}. 

\vspace{0.025in}
\noindent \textbf{\textsc{Synchronize} procedure.} 
We measure the distance between the current round $i$ of compute node $c$ and the current round of its neighbors in the history, denoted as $d_c$. If the value of $d_c$ exceeds the predefined delay threshold $d$, the training process will be postponed.
Moreover, we can get out of the \texttt{while} loop of the $\textsc{Synchronize}$ procedure at line $26$ because at some moments the compute node will broadcast to or receive a new sum of gradient 
from its neighbors. This will update the communication history $\mathcal{H}$ and eventually the delay checkpoint is satisfied.
As soon as the wait loop is exited, we know that all local gradient computations occur when $\textsc{Synchronize}(\mathcal{H}, d, i)$ is false which reflects that these gradient computations correspond to delays that are permissible.


In conclusion, the proposed \texttt{AET-SGD} framework inherits the Hogwild! algorithm. Specifically, our framework works as \textit{two Hogwild! update} phases: \textit{the inner Hogwild! update} is the actual local SGD recursions on each compute node while \textit{the outer Hogwild! update} is when the compute node updates the exchanged sum of gradient $U_{i,c}$ from its neighbors to its current local model $w$. Besides, by tracking the communication rounds, we intentionally keep the framework consistent with the delay function $\tau$ by introducing the delay checkpoint. The detailed mapping (of \texttt{AET-SGD} to Hogwild!) can be found in the next subsection.

In practice, it is hard to exactly measure and keep the network delay strictly following the delay $\tau$ (in iteration unit), but we do know there exists a threshold $d$ (in round unit), which satisfies the delay property. Hence, we can use a grid search method to figure out the 'best' value of threshold $d$.

\subsection{From Hogwild! to \texttt{AET-SGD}}
\label{sec:mapping}

In this subsection, we introduce a (virtual) timeline mapping $\rho$ to explain why \texttt{AET-SGD} can work well with linearly increasing sample sequence and be consistent with the network delay $\tau$.
Our mapping consists of two parts. First, we show that \texttt{AET-SGD} satisfies the setting of Hogwild!~\citep{Hogwild, nguyen2018new}. We then adapt the delay property $\tau$ into our algorithm.  

\vspace{0.025in}

\begin{prop} \label{prop:hogwild}
AET-SGD satisfies the setting of model aggregation of Hogwild! algorithm.
\end{prop}

\noindent \textbf{Proof.} 
The clients in the distributed computation apply recursion (\ref{eqwM2a}). We 
label each recursion with an iteration count $t$; this can then be used to compute with which delay function the labeled sequence $\{w_t \}$ is consistent to.
In order to find an ordering based on $t$ we first define a mapping $\rho$ from the annotated labels $(c,i,h)$ in \Call{MainComputeNode}{} to $t$~\citep{van2020hogwild}:
%
$$\rho(c,i,h)= \left(\sum_{l < i} s_{l} \right ) + \min\{ t' \ : \ h=|\{ t\leq t' \ : \ a(i,t)=c\}|\} ,$$ 
where the sample size sequence $\{s_i\}$ and labelling function $a(\cdot,\cdot)$ are defined in the \Call{Setup}{} procedure.
Note that given $t$, we can compute $i$ as the largest index for which $\sum_{l<i} s_{l} \leq t$, compute $t'=t- \sum_{l<i} s_{l}$ and $c=a(i,t')$, and compute $h=|\{ t\leq t' \ : \ a(i,t)=c\}|$. This procedure inverts $\rho$ since $\rho(c,i,h)=t$, and hence, $\rho$ is bijective.
From the \Call{Receive}{} procedure, we infer that the model ${w}_{k,c}$ (i.e., local model from client $c$ at round $k$) includes all the aggregate updates $U_{i,c}$ 
for $i< k$ and $c\in \{1,\ldots n\}$. In other words, there exists one point $t$, where each compute node sends or receives the gradient update $U$ to or from other compute nodes directly and indirectly, and then aggregates this update into its model. For example, we have the simple undirected network topology as \texttt{Node$_1$-Node$_2$-Node$_3$}, where \texttt{Node$_1$} and \texttt{Node$_3$} have no direct link, but \texttt{Node$_1$} can receive the gradient update of \texttt{Node$_3$} implicitly through the gradient update of \texttt{Node$_2$} and vice versa. In other words, the gradient update from \texttt{Node$_2$} implicitly includes all the aggregated updates of \texttt{Node$_1$} and \texttt{Node$_3$}. \qed



\vspace{0.025in}

\begin{prop} \label{prop:delay_property}
AET-SGD is consistent to the delay $\tau$. 
\end{prop}

\noindent \textbf{Proof.} Let $\rho(c,i,h)=t$. Notice that 
$$t= \rho(c,i,h) \leq s_0+ \ldots + s_i -(s_{i,c}-h)-1=t_{glob}$$ 
In the \Call{MainComputeNode}{} procedure, we wait as long as $\tau(t_{glob})= t_{delay}$ (in iteration unit) and this constraint is translated into $d_c > d$ (in round), where $d_c$ is the maximum distance between the current local round of the compute node and its neighbors, and the chosen network delay $d$. This means that the local SGD iteration counter $h$ will not further increase (when the Algorithm~\ref{alg:client_adaptive} enters the \texttt{while} loop at line $26$) until the delay checkpoint is satisfied. 
We conclude that the local model ${w}_{k,c}$ can include all gradient updates that correspond to gradient computations up to iteration count $t_{glob}-t_{delay}$. This includes gradient computations up to iteration count $t-\tau(t)$. Therefore, the computed sequence $\{w_t\}$ is consistent with delay function $\tau$. The realization of the delay $\tau$ is translated to procedure $\textsc{Synchronize}(\mathcal{H}, d, i)$ which enables or disables the SGD computations at the compute nodes.


Finally, from the mapping procedure above, we can confirm that \texttt{AET-SGD} follows the definition of Hogwild! algorithm~\citep{Hogwild, van2020hogwild} and this setting is 
consistent to the delay function $\tau$. \qed

\begin{remark}
Our AET-SGD framework is motivated by the setting of Hogwild! algorithm, which has the same convergence rate of SGD and is consistent to the network delay $\tau$. It means that AET-SGD will converge as the traditional SGD or the Howild! algorithm.
\end{remark}

\section{Experiments}
\label{sec:experiment}

We now present our 
experimental studies to verify the performance improvement of 
\texttt{AET-SGD}, compared with state-of-the-art methods. The experiments are conducted on {\em the ring network topology}, using representative data sets for image classification applications. Please refer to the Appendix for simulation details and complete parameter settings \footnote{The simulation source code can be found at \url{https://github.com/mlconference/AET-SGD.git}}.

\textbf{Objective function}. We conduct the experiments 
for non-convex problems, especially for image classification applications. We use a linearly increasing sample sequence $s_i = a \cdot i^p + b$, where $p=1$ and $a,b \geq 0$. For simplicity, we choose a diminishing step size scheme $\frac{\eta_0}{1 + \beta \sqrt{t}}$, where $\eta_0$ is an initial step size, $t$ is the current iteration so far and a constant $\beta \geq 0$. Also, the asynchronous SGD simulation is mainly conducted with $d=1$.

\textbf{System delay}. In order to simulate the heterogeneous environment among the compute nodes, we randomly generate the computation delay from $0.1$ to $1$ms for each SGD iteration and the network delay from $0.1$ to $1.5$ms for transmitting the sum of gradients.

\begin{figure}[ht!]
\captionsetup[subfloat]{labelformat=empty}
\vspace{-.25cm}
  \centering
  \subfloat[]{\includegraphics[width=0.5\linewidth, keepaspectratio]{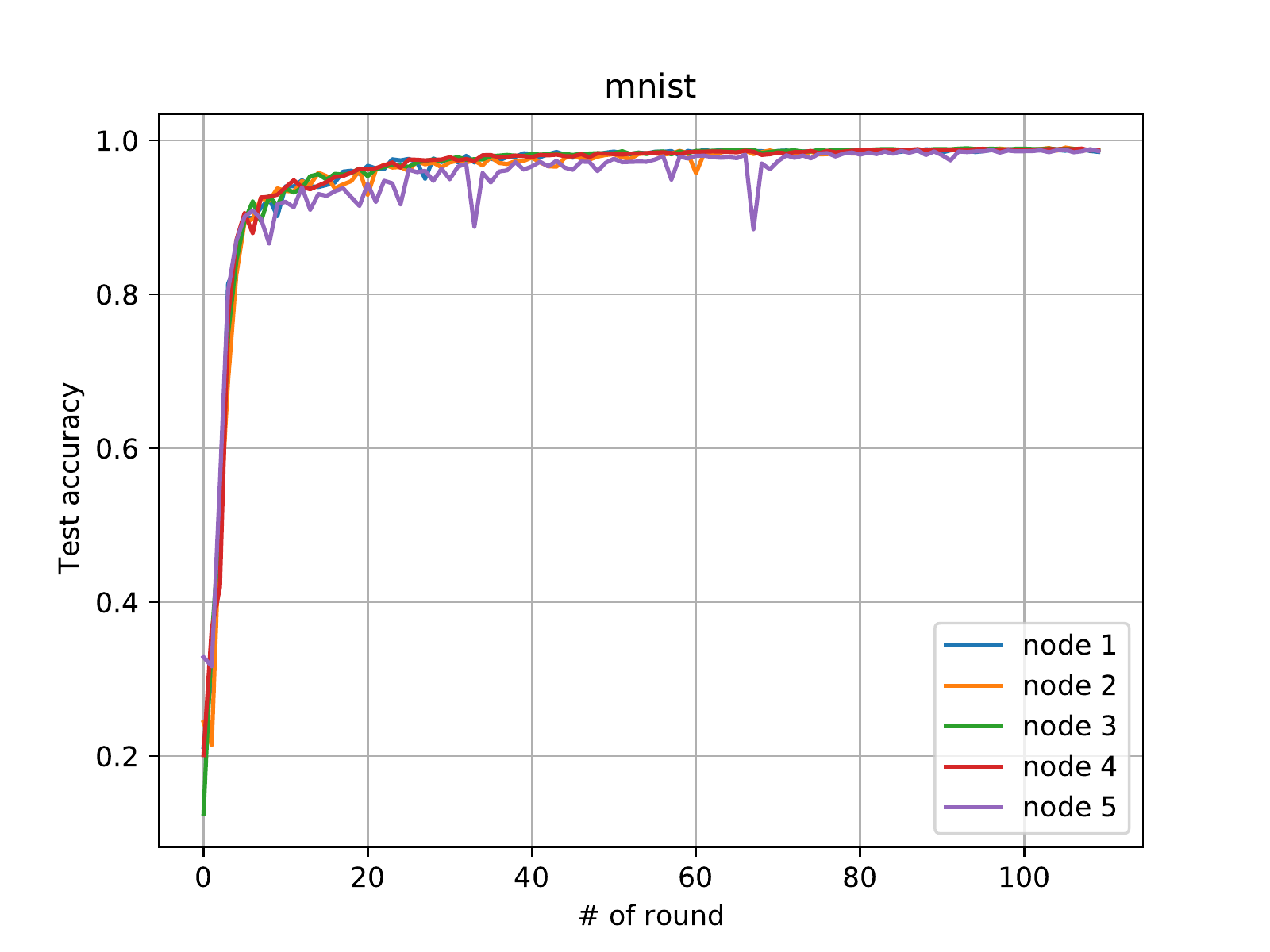}\label{fig:node_performance_mnist_main}}
  \hfill
  \subfloat[]{\includegraphics[width=0.45\linewidth,keepaspectratio]{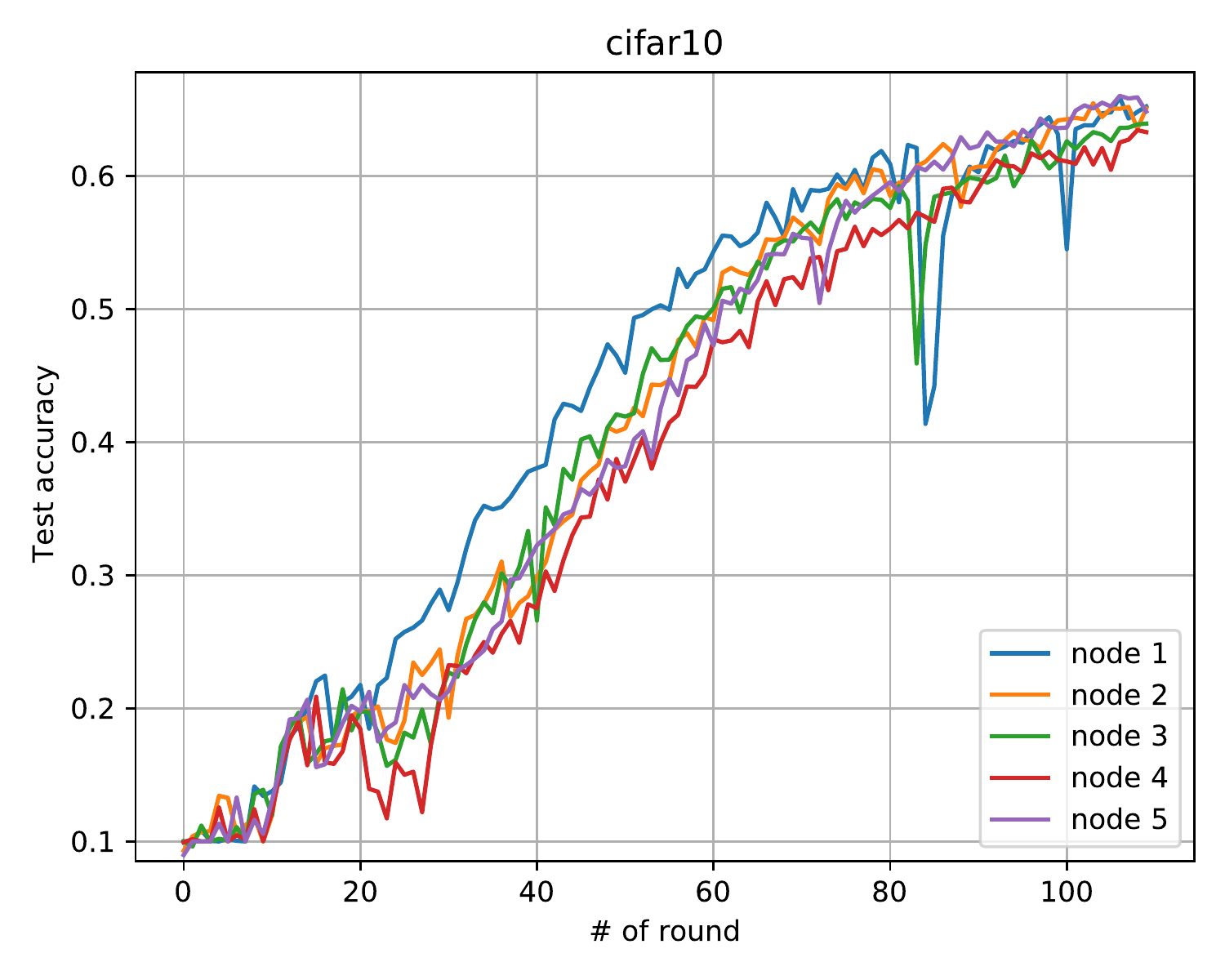} \label{fig:node_performance_cifar10_main} }
  \vspace{-0.25in}
  \caption{Test accuracy of multiple compute nodes using our proposed \texttt{AET-SGD} framework.}
  \label{fig:node_performance_main}
\end{figure}

\textbf{Convergence of \texttt{AET-SGD}}. Figure~\ref{fig:node_performance_main} illustrates the convergence rate of \texttt{AET-SGD} for MNIST and CIFAR$10$ data sets with {\em linearly increasing sample sequence}. Here, all the compute nodes joining the training process converge to nearly the same solution at the end. This confirms that \texttt{AET-SGD} can converge to a decent accuracy for all compute nodes. Experimental results for other data sets can be found in the Appendix. 
For the following experiments, we choose the test accuracy from one compute node 
to represent others, to compare with other baseline methods.

\vspace{-0.25cm}
\textbf{\texttt{AET-SGD} vs. Baseline Event-triggered SGD}. The goal of this experiment is to show that \texttt{AET-SGD} can significantly reduce the communication cost (e.g., the number of required communication rounds), compared with the baseline event-triggered SGD. For a fair comparison, two methods use the same number of iteration $K$ and start from the same initial step size $\eta_0=0.01$ (see the Appendix for details). As can be seen from Table~\ref{tbl:speedup_main}, our \texttt{AET-SGD} framework can obtain the same level of test accuracy, compared to the baseline event-triggered SGD while reducing the number of communication rounds. For example, the ratio of communication reduction can reach to $120x$ for the FashionMNIST data set. 


\begin{table}[ht!]
\begin{center}
\scalebox{0.85}{
\begin{threeparttable}
\begin{tabular}{|c|c|c|c|}
\hline
Data set & Event-triggered SGD & AET-SGD & Reduction \\
\hline
\hline
MNIST  &  $4980$ $^{\dag}$ / $0.9899$      &   $110$ / $0.9892$      &    $\approx 44$ 
\\ \hline
FashionMNIST &    $13180$ / $0.8876$       &   $110$ / $0.8868$      & $\approx 120$ 
\\ \hline
KMNIST &   $7008$ / $0.9377$    &     $110$ / $0.9346$   &  $\approx 64$
\\ \hline
CIFAR10 &   $8112$ / $0.6590$    &     $110$ / $0.6538$   &  $\approx 74$
\\ \hline
\end{tabular}
 \begin{tablenotes}
      \item {\footnotesize $\dag$ This pair of values represents the total number of communication rounds needed to finish $K$ iteration in each compute node, and the final test accuracy.}
  \end{tablenotes}
  \end{threeparttable}
}
\end{center}
\vskip -0.25cm
\caption{Comparison of communication round and test accuracy between the baseline Event-triggered SGD and \texttt{AET-SGD}}
\label{tbl:speedup_main}
\end{table}

\textbf{Scalability of \texttt{AET-SGD}}. Table~\ref{tbl:aet_diff_node_main} shows the final test accuracy and the training time of \texttt{AET-SGD} with various number of compute nodes. Here, when the number of compute node $n$ increases, the training time starts to decrease. However, when $n=15$, the training duration nearly keep unchanged (i.e., there is a speedup saturation in the distributed learning framework).
This can be explained by the effect of delay checkpoint with $d=1$, which forces the compute nodes to wait for others and eventually increase the training time. We can clearly observer the impact of the asynchronous round $d$ in Table~\ref{tbl:asymchrnous_round_main}, where we can achieve the training speedup ratio about $3.14x$ (compared to $d=0$) for the CIFAR$10$ data set when the asynchronous round $d=7$ without test accuracy degradation. Note that $d=7$ indicates a large asynchronous level among the compute nodes, caused by the straggler nodes.
This confirms that our framework can scale to a large number of compute nodes (i.e., enlarge the level of speedup saturation) if we choose the appropriate asynchronous round (e.g., $d \leq 7, n \geq 10$).
More results with other data sets can be found in the Appendix.

\vspace{-0.15cm}
\begin{table}[ht!]
\vskip 0.25cm

\begin{center}
\scalebox{0.8}{
\begin{threeparttable}
\begin{tabular}{|c|c|c|c|c|}
\hline
Data set & \# of node & Accuracy & Duration & Speedup \\ 
\hline
\hline
\multirow{5}{*}{MNIST} & $1$  &   $0.9883$   &     $182$  & $1.0$
\\ 
\cline{2-5}
& $2$   &   $0.9906$       &    $132$      & $1.38$
\\ 
\cline{2-5}
& $5$   &  $0.9881$      &     $121$   & $1.50$
\\ 
\cline{2-5}
& $10$        &   $0.9850$       &     $110$ & $1.65$
\\ 
\cline{2-5}
& $15$        &  $0.9837$       &     $109$   & $1.67$
\\ 
\hline
\hline
\multirow{5}{*}{CIFAR10} & $1$ & $0.6657$ & $3061$  &   $1.0$
\\ 
\cline{2-5}
& $2$   &   $0.6634$       &    $2414$     & $1.27$
\\ 
\cline{2-5}
& $5$   &  $0.6446$      &     $2172$   & $1.41$
\\ 
\cline{2-5}
& $10$        &   $0.6433$       &     $2053$ & $1.50$
\\ 
\cline{2-5}
& $15$        &  $0.6256$       &     $1859$   &  $1.65$
\\ 
\hline
\end{tabular}
 
      
  \end{threeparttable}
}
\end{center}

\vskip -0.15cm
\caption{Test accuracy of \texttt{AET-SGD} with different number of compute nodes $n$ and the total number of iteration $K_{total}=60,000$ for all compute nodes, i.e., $K= \ceil{ K_{total/n} }$ iterations for each compute node, $d=1$}
\label{tbl:aet_diff_node_main}

\end{table}

\vskip -0.5cm

\begin{table}[ht!]
\begin{center}
\scalebox{0.825}{
\begin{tabular}{|c|c|c|c|}
\hline
Data set & Async. round $d$  & Accuracy & Duration \\ \hline
\hline
\multirow{6}{*}{MNIST}   & $0$  &  $0.9886$        &      $179$    
\\ \cline{2-4} 
                         & $1$  &  $0.9881$        &      $121$
\\ \cline{2-4}
                         & $2$  &  $0.9888$        &    $112$      
\\ \cline{2-4}
                         & $5$  &  $0.9812$        &    $78$      
\\ \cline{2-4}
                         & 7  &    $0.9768$      &      $62$    
\\ \cline{2-4}
                         & 10 &    $0.9612$      &      $60$    
\\ \hline
\hline

\multirow{5}{*}{CIFAR10} & $0$  &    $0.6486$      &  $2792$
\\ \cline{2-4} 
                        & $1$  &    $0.6446$      &  $2172$  
\\ \cline{2-4} 
                         & $2$  &   $0.6432$       &    $1786$      
\\ \cline{2-4} 
                         & $5$  &   $0.6398$       &    $1265$      
\\ \cline{2-4} 
                         & $7$  &   $0.6331$       &    $890$      
\\ \cline{2-4} 
                         & $10$ &   $0.6019$       &    $868$      
\\ \hline
\end{tabular}
}
\end{center}
\vskip -0.25cm
\caption{Test accuracy of \texttt{AET-SGD} with different values of asynchronous round $d$ and $n=5$}
\label{tbl:asymchrnous_round_main}
\end{table}

\section{Conclusion and Future Work}
\label{sec:conclusion}
In this paper, we introduce an asynchronous event-triggered SGD framework with linearly increasing sample sequence, called \texttt{AET-SGD}.  \texttt{AET-SGD} is designed to be consistent with {the computation and network delay in heterogeneous environment} and to provide significant communication cost reduction, comparing to the baseline event-triggered SGD and other constant local SGD settings.  
We performed extensive experiments on different levels of asynchrony and our results confirm that \texttt{AET-SGD} can tolerate a high number of asynchronous rounds while keeping a good performance, e.g., the test accuracy.
The empirical evidences also confirm the communication efficiency of \texttt{AET-SGD} which can achieve a significant communication cost reduction from $48$x to $120$x, compared to the state of the art, using representative data sets. As the future work, we plan to apply model quantization or model compression methods to reduce the required bandwidth for exchanging gradient information among the compute nodes, especially for wireless devices with limited 
bandwidth and computation capacity.

\bibliography{main}

\begin{thebibliography}{39}
\providecommand{\natexlab}[1]{#1}

\bibitem[{Bonawitz et~al.(2019)Bonawitz, Eichner, Grieskamp, Huba, Ingerman,
  Ivanov, Kiddon, Konecny, Mazzocchi, McMahan et~al.}]{bonawitz2019towards}
Bonawitz, K.; Eichner, H.; Grieskamp, W.; Huba, D.; Ingerman, A.; Ivanov, V.;
  Kiddon, C.; Konecny, J.; Mazzocchi, S.; McMahan, H.~B.; et~al. 2019.
\newblock Towards federated learning at scale: System design.
\newblock \emph{Proceedings of the $2$nd SysML Conference}.

\bibitem[{Bottou, Curtis, and Nocedal(2018)}]{bottou2018optimization}
Bottou, L.; Curtis, F.~E.; and Nocedal, J. 2018.
\newblock Optimization methods for large-scale machine learning.
\newblock \emph{Siam Review}, 60(2): 223--311.

\bibitem[{Chen et~al.(2016)Chen, Monga, Bengio, and Jozefowicz}]{jianmin}
Chen, J.; Monga, R.; Bengio, S.; and Jozefowicz, R. 2016.
\newblock Revisiting distributed synchronous SGD.
\newblock \emph{ICLR Workshop Track}.

\bibitem[{Chen, Sun, and Jin(2019)}]{yang2019}
Chen, Y.; Sun, X.; and Jin, Y. 2019.
\newblock Communication-efficient federated deep learning with layerwise
  asynchronous model update and temporally weighted aggregation.
\newblock \emph{IEEE transactions on neural networks and learning systems},
  31(10): 4229--4238.

\bibitem[{Clanuwat et~al.(2018)Clanuwat, Bober-Irizar, Kitamoto, Lamb,
  Yamamoto, and Ha}]{clanuwat2018deep}
Clanuwat, T.; Bober-Irizar, M.; Kitamoto, A.; Lamb, A.; Yamamoto, K.; and Ha,
  D. 2018.
\newblock Deep Learning for Classical Japanese Literature.

\bibitem[{De~Sa et~al.(2015)De~Sa, Zhang, Olukotun, and
  R{\'e}}]{DeSaZhangOlukotunEtAl2015}
De~Sa, C.~M.; Zhang, C.; Olukotun, K.; and R{\'e}, C. 2015.
\newblock {Taming the wild: A unified analysis of hogwild-style algorithms}.
\newblock In \emph{NIPS}, 2674--2682.

\bibitem[{Dean et~al.(2012)Dean, Corrado, Monga, Chen, Devin, Le, Mao, Ranzato,
  Senior, Tucker, Yang, and Ng}]{dean2012large}
Dean, J.; Corrado, G.~S.; Monga, R.; Chen, K.; Devin, M.; Le, Q.~V.; Mao,
  M.~Z.; Ranzato, M.; Senior, A.; Tucker, P.; Yang, K.; and Ng, A.~Y. 2012.
\newblock Large Scale Distributed Deep Networks.
\newblock In \emph{NIPS}.

\bibitem[{George and Gurram(2020)}]{george2019distributed}
George, J.; and Gurram, P. 2020.
\newblock Distributed Deep Learning with Event-Triggered Communication.
\newblock In \emph{The Thirty-Fourth AAAI Conference on Artificial
  Intelligence}, 7169--7178.

\bibitem[{Haddadpour et~al.(2019)Haddadpour, Kamani, Mahdavi, and
  Cadambe}]{haddadpour2019local}
Haddadpour, F.; Kamani, M.~M.; Mahdavi, M.; and Cadambe, V. 2019.
\newblock Local SGD with Periodic Averaging: Tighter Analysis and Adaptive
  Synchronization.
\newblock \emph{Advances in Neural Information Processing Systems}, 32:
  11082--11094.

\bibitem[{Hsieh et~al.(2017)Hsieh, Harlap, Vijaykumar, Konomis, Ganger,
  Gibbons, and Mutlu}]{hsieh2017gaia}
Hsieh, K.; Harlap, A.; Vijaykumar, N.; Konomis, D.; Ganger, G.~R.; Gibbons,
  P.~B.; and Mutlu, O. 2017.
\newblock Gaia: Geo-distributed machine learning approaching $\{$LAN$\}$
  speeds.
\newblock In \emph{14th $\{$USENIX$\}$ Symposium on Networked Systems Design
  and Implementation ($\{$NSDI$\}$ 17)}, 629--647.

\bibitem[{Kone{\v{c}}n{\`y} et~al.(2016)Kone{\v{c}}n{\`y}, McMahan, Ramage, and
  Richt{\'a}rik}]{konevcny2016federated}
Kone{\v{c}}n{\`y}, J.; McMahan, H.~B.; Ramage, D.; and Richt{\'a}rik, P. 2016.
\newblock Federated optimization: Distributed machine learning for on-device
  intelligence.
\newblock \emph{arXiv preprint arXiv:1610.02527}.

\bibitem[{Krizhevsky, Hinton et~al.(2009)}]{krizhevsky2009learning}
Krizhevsky, A.; Hinton, G.; et~al. 2009.
\newblock Learning multiple layers of features from tiny images.

\bibitem[{Krizhevsky, Sutskever, and Hinton(2012)}]{krizhevsky2012imagenet}
Krizhevsky, A.; Sutskever, I.; and Hinton, G.~E. 2012.
\newblock Imagenet classification with deep convolutional neural networks.
\newblock \emph{Advances in neural information processing systems}, 25:
  1097--1105.

\bibitem[{Leblond, Pedregosa, and Lacoste-Julien(2018)}]{Leblond2018}
Leblond, R.; Pedregosa, F.; and Lacoste-Julien, S. 2018.
\newblock Improved asynchronous parallel optimization analysis for stochastic
  incremental methods.
\newblock \emph{JMLR}, 19(1): 3140--3207.

\bibitem[{LeCun et~al.(1998)LeCun, Bottou, Bengio, and
  Haffner}]{lecun1998gradient}
LeCun, Y.; Bottou, L.; Bengio, Y.; and Haffner, P. 1998.
\newblock Gradient-based learning applied to document recognition.
\newblock \emph{Proceedings of the IEEE}, 86(11): 2278--2324.

\bibitem[{LeCun and Cortes(2010)}]{lecun-mnisthandwrittendigit-2010}
LeCun, Y.; and Cortes, C. 2010.
\newblock {MNIST} handwritten digit database.

\bibitem[{Li et~al.(2019)Li, Sahu, Zaheer, Sanjabi, Talwalkar, and
  Smith}]{tianli2019hetero}
Li, T.; Sahu, A.~K.; Zaheer, M.; Sanjabi, M.; Talwalkar, A.; and Smith, V.
  2019.
\newblock Federated Optimization for Heterogeneous Networks.
\newblock \emph{arXiv preprint}.

\bibitem[{Lian et~al.(2015)Lian, Huang, Li, and Liu}]{lian2015asynchronous}
Lian, X.; Huang, Y.; Li, Y.; and Liu, J. 2015.
\newblock Asynchronous parallel stochastic gradient for nonconvex optimization.
\newblock In \emph{Advances in Neural Information Processing Systems},
  2737--2745.

\bibitem[{Lin et~al.(2020)Lin, Stich, Patel, and Jaggi}]{lin2018don}
Lin, T.; Stich, S.~U.; Patel, K.~K.; and Jaggi, M. 2020.
\newblock Don't Use Large Mini-Batches, Use Local SGD.
\newblock \emph{International Conference on Learning Representations (ICLR)}.

\bibitem[{Mania et~al.(2015)Mania, Pan, Papailiopoulos, Recht, Ramchandran, and
  Jordan}]{ManiaPanPapailiopoulosEtAl2015}
Mania, H.; Pan, X.; Papailiopoulos, D.; Recht, B.; Ramchandran, K.; and Jordan,
  M.~I. 2015.
\newblock {Perturbed Iterate Analysis for Asynchronous Stochastic
  Optimization}.
\newblock \emph{SIAM Journal on Optimization}, 2202--2229.

\bibitem[{McMahan et~al.(2016)McMahan, Moore, Ramage, and y~Arcas}]{mcmahan}
McMahan, H.~B.; Moore, E.; Ramage, D.; and y~Arcas, B.~A. 2016.
\newblock Federated Learning of Deep Networks using Model Averaging.
\newblock \emph{ICLR Workshop Track}.

\bibitem[{Meng et~al.(2017)Meng, Chen, Yu, Wang, Ma, and
  Liu}]{meng2017asynchronous}
Meng, Q.; Chen, W.; Yu, J.; Wang, T.; Ma, Z.-M.; and Liu, T.-Y. 2017.
\newblock Asynchronous stochastic proximal optimization algorithms with
  variance reduction.
\newblock In \emph{Thirty-First AAAI Conference on Artificial Intelligence}.

\bibitem[{Nguyen et~al.(2019)Nguyen, Nguyen, Richt{{\'a}}rik, Scheinberg,
  Tak{{\'a}}{\v{c}}, and van Dijk}]{nguyen2018new}
Nguyen, L.~M.; Nguyen, P.~H.; Richt{{\'a}}rik, P.; Scheinberg, K.;
  Tak{{\'a}}{\v{c}}, M.; and van Dijk, M. 2019.
\newblock New Convergence Aspects of Stochastic Gradient Algorithms.
\newblock \emph{Journal of Machine Learning Research}, 20(176): 1--49.

\bibitem[{Nguyen et~al.(2018)Nguyen, Nguyen, van Dijk, Richt{\'{a}}rik,
  Scheinberg, and Tak{\'{a}}c}]{nguyen2018sgd}
Nguyen, L.~M.; Nguyen, P.~H.; van Dijk, M.; Richt{\'{a}}rik, P.; Scheinberg,
  K.; and Tak{\'{a}}c, M. 2018.
\newblock {SGD} and Hogwild! Convergence Without the Bounded Gradients
  Assumption.
\newblock In \emph{Proceedings of the 35th International Conference on Machine
  Learning, {ICML} 2018}, 3747--3755.

\bibitem[{Nguyen et~al.(2021)Nguyen, Nguyen, Nguyen, Tran-Dinh, Nguyen, and
  Dijk}]{van2020hogwild}
Nguyen, N.; Nguyen, T.; Nguyen, P.~H.; Tran-Dinh, Q.; Nguyen, L.; and Dijk, M.
  2021.
\newblock Hogwild! over Distributed Local Data Sets with Linearly Increasing
  Mini-Batch Sizes.
\newblock In \emph{International Conference on Artificial Intelligence and
  Statistics}, 1207--1215. PMLR.

\bibitem[{Recht et~al.(2011)Recht, Re, Wright, and Niu}]{Hogwild}
Recht, B.; Re, C.; Wright, S.; and Niu, F. 2011.
\newblock Hogwild: A lock-free approach to parallelizing stochastic gradient
  descent.
\newblock In \emph{Advances in neural information processing systems},
  693--701.

\bibitem[{Robbins and Monro(1951)}]{robbins1951stochastic}
Robbins, H.; and Monro, S. 1951.
\newblock A stochastic approximation method.
\newblock \emph{The annals of mathematical statistics}, 400--407.

\bibitem[{Roux, Schmidt, and Bach(2012)}]{roux2012stochastic}
Roux, N.; Schmidt, M.; and Bach, F. 2012.
\newblock A Stochastic Gradient Method with an Exponential Convergence Rate for
  Finite Training Sets.
\newblock In Pereira, F.; Burges, C. J.~C.; Bottou, L.; and Weinberger, K.~Q.,
  eds., \emph{Advances in Neural Information Processing Systems}, volume~25.
  Curran Associates, Inc.

\bibitem[{Sheng et~al.(2020)Sheng, Wang, Jin, Yan, Li, Chang, Wang, and
  Zha}]{sheng2020learning}
Sheng, J.; Wang, X.; Jin, B.; Yan, J.; Li, W.; Chang, T.-H.; Wang, J.; and Zha,
  H. 2020.
\newblock Learning Structured Communication for Multi-agent Reinforcement
  Learning.
\newblock \emph{arXiv preprint arXiv:2002.04235}.

\bibitem[{Shi et~al.(2019)Shi, Wang, Zhao, Tang, Wang, Huang, and
  Chu}]{shi2019distributed}
Shi, S.; Wang, Q.; Zhao, K.; Tang, Z.; Wang, Y.; Huang, X.; and Chu, X. 2019.
\newblock A distributed synchronous SGD algorithm with global top-k
  sparsification for low bandwidth networks.
\newblock In \emph{2019 IEEE 39th International Conference on Distributed
  Computing Systems (ICDCS)}, 2238--2247. IEEE.

\bibitem[{Stich(2019)}]{stich2018local}
Stich, S.~U. 2019.
\newblock Local SGD converges fast and communicates little.
\newblock \emph{International Conference on Learning Representations (ICLR)}.

\bibitem[{Wang, Wang, and Li(2019)}]{luping}
Wang, L.; Wang, W.; and Li, B. 2019.
\newblock CMFL: Mitigating Communication Overhead for Federated Learning.
\newblock \emph{IEEE International Conference on Distributed Computing
  Systems.}

\bibitem[{Xiao, Rasul, and Vollgraf(2017)}]{xiao2017fashion}
Xiao, H.; Rasul, K.; and Vollgraf, R. 2017.
\newblock Fashion-mnist: a novel image dataset for benchmarking machine
  learning algorithms.
\newblock \emph{arXiv preprint arXiv:1708.07747}.

\bibitem[{Xie, Koyejo, and Gupta(2019)}]{cong2019}
Xie, C.; Koyejo, S.; and Gupta, I. 2019.
\newblock Asynchronous Federated Optimization.
\newblock \emph{arXiv preprint}.

\bibitem[{Yin et~al.(2018)Yin, Pananjady, Lam, Papailiopoulos, Ramchandran, and
  Bartlett}]{yin2018gradient}
Yin, D.; Pananjady, A.; Lam, M.; Papailiopoulos, D.; Ramchandran, K.; and
  Bartlett, P. 2018.
\newblock Gradient diversity: a key ingredient for scalable distributed
  learning.
\newblock In \emph{International Conference on Artificial Intelligence and
  Statistics}, 1998--2007. PMLR.

\bibitem[{You, Gitman, and Ginsburg(2017)}]{you2017scaling}
You, Y.; Gitman, I.; and Ginsburg, B. 2017.
\newblock Scaling sgd batch size to 32k for imagenet training.
\newblock \emph{arXiv preprint arXiv:1708.03888}, 6: 12.

\bibitem[{You et~al.(2018)You, Zhang, Hsieh, Demmel, and
  Keutzer}]{you2018imagenet}
You, Y.; Zhang, Z.; Hsieh, C.-J.; Demmel, J.; and Keutzer, K. 2018.
\newblock Imagenet training in minutes.
\newblock In \emph{Proceedings of the 47th International Conference on Parallel
  Processing}, 1--10.

\bibitem[{Zheng et~al.(2017)Zheng, Meng, Wang, Chen, Yu, Ma, and
  Liu}]{zheng2017asynchronous}
Zheng, S.; Meng, Q.; Wang, T.; Chen, W.; Yu, N.; Ma, Z.-M.; and Liu, T.-Y.
  2017.
\newblock Asynchronous stochastic gradient descent with delay compensation.
\newblock In \emph{Proceedings of the 34th International Conference on Machine
  Learning-Volume 70}, 4120--4129. JMLR. org.

\bibitem[{Zinkevich, Langford, and Smola(2009)}]{zinkevich2009slow}
Zinkevich, M.; Langford, J.; and Smola, A.~J. 2009.
\newblock Slow learners are fast.
\newblock In \emph{Advances in neural information processing systems},
  2331--2339.

\end{thebibliography}

\section{Appendix: Details of the Experimental Setting and Results}
\label{sec:experiment_suppl}

In this Appendix, we present our results from comprehensive experimental studies to verify the performance improvement of the proposed \texttt{AET-SGD} framework compared with state-of-the-art methods. The experiments are conducted on {the ring network topology}, using representative data sets for image classification applications.




\subsection{Experiment settings}
\label{sec:exp_setting}



\vspace{0.025in}
\noindent \textbf{Simulation environment.} For simulating the \texttt{AET-SGD} and other baseline frameworks, we use multiple threads where each thread represents one client joining the training process. The experiments are conducted on Linux-64bit OS, 
$32$Gb RAM and $2$ NVIDIA GPUs. We mainly choose the ring network topology with $5$ compute nodes as shown in Figure~\ref{fig:ring_topology}.  Moreover, we randomly add the delay on both computation and communication process to simulate the effect of straggler nodes.

\vspace{0.025in}
\noindent \textbf{Objective function.} Our experiments mainly focus on non-convex problems (deep neural networks). For simplicity, we choose the
neural network--LeNet~\citep{lecun1998gradient} and AlexNet~\citep{krizhevsky2012imagenet} for image classification applications.

\vspace{0.025in}
\noindent \textbf{Data sets.} We choose representative data sets, such as MNIST~\citep{lecun-mnisthandwrittendigit-2010}, FashionMNIST~\citep{xiao2017fashion}, KMNIST~\citep{clanuwat2018deep} and CIFAR10~\citep{krizhevsky2009learning} for our experiments. We use Pytorch library to implement our proposed framework.

\vspace{0.025in}
\noindent \textbf{Baselines.} The baseline algorithm is the distributed event-triggered SGD~\citep{george2019distributed}. For fair comparison, we use the same number of compute nodes, diminishing step size scheme, and number of total iterations to conduct the experiments. We also compare \texttt{AET-SGD} with the constant local SGD framework~\citep{stich2018local}.  

\vspace{0.025in}
\noindent \textbf{Hyper-parameters selection.} We use the same initial step size $\eta_0 = 0.01$ for both the baseline event-triggered SGD~\citep{george2019distributed} and \texttt{AET-SGD}. Here, the baseline algorithm uses the step size $\alpha_t = \frac{\eta_0}{\epsilon \cdot t + 1}$ and $\beta_t = \frac{2.252 \cdot \eta_0}{(\epsilon \cdot t + 1)^{1/10}}$, where $\epsilon = 10^{-5}$ and the event-triggered threshold is $\eta_t \cdot v_0$, where $v_0 = 0.2 \cdot N_{p}$, $N_{p}$ is the total number of parameters in each neural network and $\eta_t$ is the current step size. In our proposed algorithm, we use the diminishing step size scheme $\frac{\eta_0}{1 + \beta \cdot \sqrt{t}}$.
Parameter $\eta_0$ is the initial step size and $\beta=0.01$, which we compute by performing a systematic grid search for $\beta$ (i.e., we select the $\beta$ giving 'best' convergence). Moreover, the general linear increasing sample sequence is chosen as $s_i \sim \mathcal{O}(i)$, where $i$ is the current communication round. {\em In practice, there are various linear increasing sample sequences that can work and achieve a good performance. In this work, we pick a simple one to demonstrate our proposed algorithm.}

To make the analysis simple, we mostly consider the asynchronous behaviour with $d=1$ (i.e., each compute node is allowed to run faster than its neighbors for at most $1$ communication round) and unbiased data sets. Moreover, the probability $p_c$ is set to be the same for all compute nodes, i.e., each compute node has the uniform distribution of the whole data set. The basic parameter setting is summarized in Table~\ref{tbl:basic_param_setting}.

\begin{table}[ht!]
\begin{center}
\scalebox{0.85}{
\begin{threeparttable}
\begin{tabular}{|c|c|}
\hline
Parameter & Value \\ \hline \hline
Initial stepsize $\bar{\eta}_0$  & $0.01$
\\ \hline
Stepsize  scheme & $\textstyle {\eta}_i = \textstyle \frac{\bar{\eta}_0}{1 + \beta \cdot \sqrt{t}}, \beta = 0.01$
\\ \hline
\# of data points $N_c$ &    $50,000$ or $60,000$
\\ \hline
\# of iterations $K$ &    $60,000$
\\ \hline
Linear sample size $s_i$ &    $s_i = a \cdot i^p + b$ ($a=10, p=1, b=0$)
\\ \hline
Constant sample size $s$ & $\{10, 50, 100, 200, 500, 700, 1,000 \}$
\\ \hline
Data set & \texttt{MNIST, FashionMNIST} \\ & \texttt{KMNIST, CIFAR10}
\\ \hline
 \# of nodes $n$ & $5$
\\ \hline
Asynchronous round $d$ & $1$
\\ \hline
Network topology & ring topology \\
\hline
\end{tabular}
  \end{threeparttable}
}
\end{center}
\caption{Summary of the default parameter setting}
\label{tbl:basic_param_setting}
\end{table}

For performance evaluation, we define the test accuracy as the the fraction of samples from a test data set that get accurately labeled by the classifier (as a result of training on a training data set by minimizing a corresponding objective function). 

{\em In addition, we randomly generate the computation delay from $0.1$ to $1$ms for each SGD iteration and the network delay from $0.1$ to $1.5$ms for transmitting the sum of gradients.}


\subsection{Convergence of \texttt{AET-SGD}}




The goal of this set of experiments is to verify the convergence of \texttt{AET-SGD}. Our expectation is that all compute nodes joining the training process will achieve the same (final) test accuracy. To conduct the experiment, we set the total number of iteration to be $K=60,000$ for each compute node. The network topology is set to be the ring network with $5$ clients with undirected connections. Also, a chosen linearly increasing sample size sequence of compute node $c$, denoted as $s_{i,c}$, is linearly chosen as $s_{i,c}=a \cdot i^p + b$, where $p=1$ and $a, b \geq 0$~\citep{van2020hogwild}. In this experiment, we simply choose the parameters as $a=10$ and $b=0$, and then the sample size sequence will be $s_{i,c} = 10 \cdot i$, where $i$ is the current communication round.  The detailed description of hyper parameters such as $\eta_0, \beta$ and the asynchronous rounds are 
summarized in Table~\ref{tbl:basic_param_setting}.
\begin{figure}[ht!]
\captionsetup[subfloat]{labelformat=empty}
  \vspace{-.60cm}
  \centering
  \subfloat[]{\includegraphics[width=0.5\linewidth, keepaspectratio]{Experiment/nodes/accuracy_1614363718_mnist_i.i.d_2_full_LOCALSGD.pdf}\label{fig:node_performance_mnist}}
  \hfill
  \subfloat[]{\includegraphics[width=0.5\linewidth,keepaspectratio]{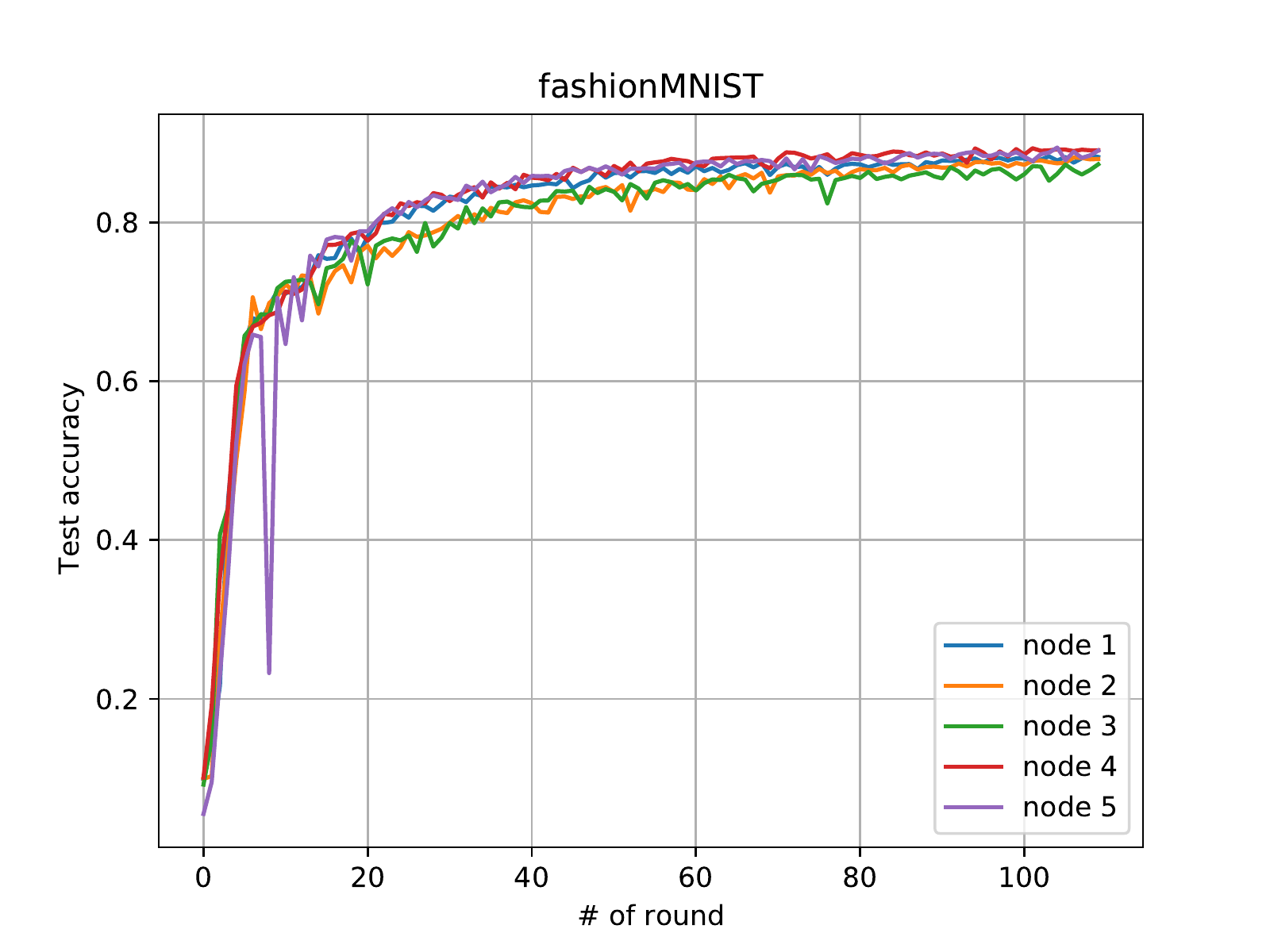}\label{fig:node_performance_fashionmnist}}
  \hfill
  \vspace{-0.25in}
  \subfloat[]{\includegraphics[width=0.5\linewidth,keepaspectratio]{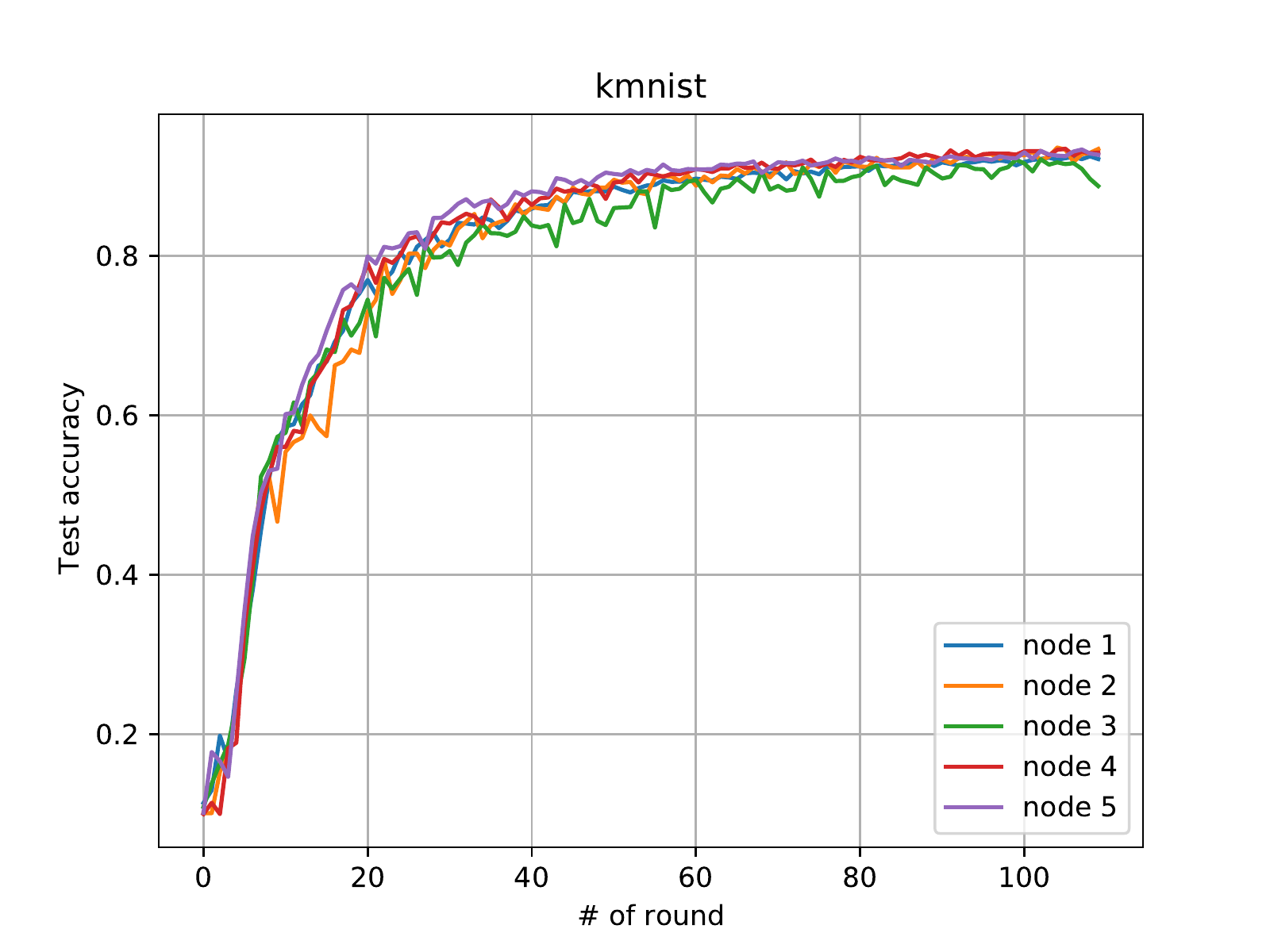}\label{fig:node_performance_kmnist}}
  \hfill
  \subfloat[]{\includegraphics[width=0.45\linewidth,keepaspectratio]{Experiment/nodes/Task0_topology_cifar10_non_convex.pdf}}
  \vspace{-0.25in}
  \caption{Test accuracy of compute nodes using our proposed \texttt{AET-SGD} framework.}
  \label{fig:node_performance}
\end{figure}

From Figure~\ref{fig:node_performance}, we can see that while the test accuracy among compute nodes might fluctuate during the training process, the final test accuracy of the compute nodes converge to nearly the same values for all the data sets. This indicates the local models using \texttt{AET-SGD} eventually converge to nearly same final solution. {\em For simplicity, we pick up the test accuracy from one specific compute node (e.g., the best one), to compare with other baseline settings in the following experiments.} Note that the difference of the test accuracy among the compute nodes is negligible, so there is a very small gap (of the test accuracy) between the best compute node and the worst compute node, which does not affect the comparison of \texttt{AET-SGD} and other methods in the following experiments.

\subsection{\texttt{AET-SGD} vs. baseline event-triggered SGD}




In this set of experiments, we compare the performance of \texttt{AET-SGD} with the baseline event-triggered SGD algorithm. We set the number of iteration $K=60,000$ for each compute node. The network topology is set to be the ring network with $5$ clients as shown in Figure~\ref{fig:ring_topology}. Also, we choose the sample size sequence as $s_{i,c} = 10 \cdot i$, where $i$ is the current communication round of the compute node $c$. Other parameter setting is the same as summarized in Table~\ref{tbl:basic_param_setting}.




\begin{table}[ht!]
\begin{center}
\scalebox{0.82}{
\begin{threeparttable}
\begin{tabular}{|c|c|c|c|}
\hline
Data set & Event-triggered SGD & AET-SGD & Reduction \\
\hline
\hline
MNIST  &  $4980$ $^{\dag}$ / $0.9899$      &   $110$ / $0.9892$      &    $\approx 44$ 
\\ \hline
FashionMNIST &    $13180$ / $0.8876$       &   $110$ / $0.8868$      & $\approx 120$ 
\\ \hline
KMNIST &   $7008$ / $0.9377$    &     $110$ / $0.9346$   &  $\approx 64$
\\ \hline
CIFAR10 &   $8112$ / $0.6590$    &     $110$ / $0.6538$   &  $\approx 74$
\\ \hline
\end{tabular}
 \begin{tablenotes}
      \item {\footnotesize $\dag$ This pair of values represents the total number of communication rounds needed to finish $K$ iteration in each compute node, and the final test accuracy.}
  \end{tablenotes}
  \end{threeparttable}
}
\end{center}

\caption{Summary of communication round and test accuracy of the baseline Event-triggered SGD and \texttt{AET-SGD}}
\label{tbl:speedup}

\end{table}

As can be observed from Table~\ref{tbl:speedup},
there is no significant difference in terms of the (final) test accuracy between the baseline algorithm and \texttt{AET-SGD} algorithm. 
This indicates that our algorithm can achieve the same level of good performance, in terms of test accuracy. 
When it comes to the number of communication rounds, \texttt{AET-SGD} shows a significant reduction, compared to the baseline algorithm. Specifically, \texttt{AET-SGD} can significantly reduce the communication cost about $44$x for the MNIST data set, $120$x for the FashionMNIST data set, $64$x for the KMNIST data set {and about $74x$ for the CIFAR$10$ data set}.
Note that we can make the baseline event-triggered threshold larger to reduce the communication rounds; however, this setting seems unfair to make the comparison, because larger threshold performs poorly in terms of the test accuracy as shown in Table~\ref{tbl:event_triggered_threshold}.




\subsection{\texttt{AET-SGD} vs. constant local SGD}
In this set of experiments, we investigate the effect of constant sample size on the test accuracy of our proposed framework. For simplicity, we conduct the experiment with $5$ compute nodes with the asynchronous behaviour $d=1$.
Moreover, we fix the number of iterations on each compute node $K=60,000$ and the diminishing step size $\eta_t = \frac{\eta_0}{1 + \beta \cdot \sqrt{t}}$. Other hyper parameters such as $\eta_0, \beta$ and the network topology are the same as in 
Table~\ref{tbl:basic_param_setting}.

As can be observed from Table~\ref{tbl:sample_size_mnist_full}, {for MNIST data set} \texttt{AET-SGD} 
can achieve the same level of accuracy, compared to other constant sample size settings. However, \texttt{AET-SGD} has significant advantages in reducing communication cost. Note that when we increase the constant sample size from $s=500$ to $1,000$ SGD recursions, the final accuracy has gradually gone down, compared to other settings. The experiments are then extended to the \texttt{FashionMNIST} \texttt{KMNIST} and {CIFAR$10$} data sets.
This confirms that \texttt{AET-SGD} with linearly increasing local SGD can achieve \textit{a good communication cost-accuracy tradeoff}, compared to other local SGD settings.

\begin{table}[ht!]
\begin{center}
\scalebox{0.9}{
\begin{threeparttable}
\begin{tabular}{|c|c|c|c|}
\hline
Data set  & Sample size  & \# of round & Accuracy \\ 
\hline
\hline
\multirow{8}{*}{MNIST} & $10$    &    $6,000$  &      $0.9905$         
\\ \cline{2-4}
& $50$    &    $1,200$   &     $0.9897$                
\\ \cline{2-4}
& $100$   &    $600$     &     $0.9894$                
\\ \cline{2-4}
& $200$   &    $300$     &     $0.9883$                
\\ \cline{2-4}
& $500$   &    $120$     &     $0.9742$                
\\ \cline{2-4}
& $700$   &    $86$     &      $0.9740$               
\\ \cline{2-4}
& $1,000$  &    $60$     &     $0.9682$           
\\ \cline{2-4}
& \texttt{Increasing}$\dag$ &  $110$  &  $0.9892$            
\\ 
\hline
\hline

\multirow{8}{*}{FashionMNIST} & $10$    &   $6,000$         &   $0.8866$                 
\\ \cline{2-4}
& $50$    &   $1,200$         &   $0.8765$               
\\ \cline{2-4}
& $100$   &   $600$           &  $0.8761$                  
\\ \cline{2-4}
& $200$   &   $300$           &   $0.8738$                 
\\ \cline{2-4}
& $500$   &   $120$           &   $0.8627$                 
\\ \cline{2-4}
& $700$   &    $86$           &   $0.8523$                 
\\ \cline{2-4}
& $1,000$  &  $60$            &   $0.8438$
\\ \cline{2-4}
& \texttt{Increasing} &   $110$ &    $0.8868$         
\\ \hline \hline

\multirow{8}{*}{KMNIST} & $10$    &    $6,000$  &        $0.9357$            
\\ \cline{2-4}
& $50$    &    $1,200$   &       $0.9288$                
\\ \cline{2-4}
& $100$   &    $600$     &       $0.9285$            
\\ \cline{2-4}
& $200$   &    $300$     &       $0.9242$                
\\ \cline{2-4}
& $500$   &    $120$     &       $0.9146$                
\\ \cline{2-4}
& $700$   &    $86$     &        $0.9035$               
\\ \cline{2-4}
& $1,000$  &     $60$     &      $0.8931$                 
\\ \cline{2-4}
& \texttt{Increasing} &  $110$  &  $0.9346$            \\ \hline \hline

\multirow{8}{*}{CIFAR$10$} & $10$    &    $6,000$  &  $0.6520$
\\ \cline{2-4}
& $50$    &    $1,200$   &       $0.6454$                
\\ \cline{2-4}
& $100$   &    $600$     &       $0.6168$            
\\ \cline{2-4}
& $200$   &    $300$     &       $0.6145$                
\\ \cline{2-4}
& $500$   &    $120$     &       $0.6047$                
\\ \cline{2-4}
& $700$   &    $86$     &        $0.6023$               
\\ \cline{2-4}
& $1,000$  &     $60$     &      $0.5963$                 
\\ \cline{2-4}
& \texttt{Increasing} &  $110$  &  $0.6585$
\\ \hline

\end{tabular}
 \begin{tablenotes}
      \item {\footnotesize $\dag$ \texttt{Increasing} means the linearly increasing sample sequences.}
  \end{tablenotes}
  \end{threeparttable}
}
\end{center}

\caption{Test accuracy of \texttt{AET-SGD} with constant and linearly increasing sample sizes}
\label{tbl:sample_size_mnist_full}

\end{table}

\subsection{\texttt{AET-SGD} with different number of iterations}
\label{sec:iteration}

We now want to understand how varying the number of iterations $K$ of each compute node while fixing other parameters, such as the number  $n$ of compute nodes  and diminishing step size sequence, affects the accuracy.
The goal of this experiment is to show that from some point onward it does not help to increase the number of iterations. The hyper parameters such as $\eta_0, \beta$ and the network topology are the same as shown in 
Table~\ref{tbl:basic_param_setting}. Note that we use \textit{the iid data sets} for each compute node in this experiment.

\begin{table}[ht!]

\begin{center}
\scalebox{0.9}{
\begin{threeparttable}
\begin{tabular}{|c|c|c|}
\hline
Data set & \# of iteration & Accuracy     
\\ \hline \hline
\multirow{8}{*}{MNIST} & 1,000   &   $0.9391$
\\ 
\cline{2-3}
& 2,000         &       $0.9581$        
\\ 
\cline{2-3}
& 5,000         &       $0.9727$        
\\ 
\cline{2-3}
&  10,000        &      $0.9807$        
\\ 
\cline{2-3}
&  20,000        &      $0.9838$       
\\ 
\cline{2-3}
& 50,000       &        $0.9887$        
\\ 
\cline{2-3}
& 75,000       &        $0.9896$        
\\ 
\cline{2-3}
& 100,000       &      $0.9899$      
\\
\hline \hline

\multirow{8}{*}{FashionMNIST} & 1,000   &       $0.7280$
\\ 
\cline{2-3}
& 2,000         &       $0.7589$        
\\ 
\cline{2-3}
& 5,000         &       $0.8013$       
\\ 
\cline{2-3}
&  10,000        &      $0.8394$        
\\ 
\cline{2-3}
&  20,000        &      $0.8661$       
\\ 
\cline{2-3}
& 50,000       &        $0.8813$        
\\ 
\cline{2-3}
& 75,000       &        $0.8863$        
\\ 
\cline{2-3}
& 100,000       &      $0.8819$      
\\
\hline \hline

\multirow{8}{*}{KMNIST} & 1,000   &       $0.6702$
\\ 
\cline{2-3}
& 2,000         &       $0.7619$        
\\ 
\cline{2-3}
& 5,000         &       $0.8579$        
\\ 
\cline{2-3}
&  10,000        &      $0.8869$        
\\ 
\cline{2-3}
&  20,000        &      $0.9153$       
\\ 
\cline{2-3}
& 50,000       &        $0.9320$        
\\ 
\cline{2-3}
& 75,000       &        $0.9344$        
\\ 
\cline{2-3}
& 100,000       &      $0.9359$      
\\
\hline \hline

\multirow{8}{*}{CIFAR10} & 1,000   &       $0.1952$
\\ 
\cline{2-3}
& 2,000         &       $0.2223$        
\\ 
\cline{2-3}
& 5,000         &       $0.2828$        
\\ 
\cline{2-3}
&  10,000        &      $0.3558$        
\\ 
\cline{2-3}
&  20,000        &      $0.5035$       
\\ 
\cline{2-3}
& 50,000       &        $0.6483$        
\\ 
\cline{2-3}
& 75,000       &        $0.6635$        
\\ 
\cline{2-3}
& 100,000       &      $0.6595$      
\\
\hline

\end{tabular}
  \end{threeparttable}
}
\end{center}

\caption{Test accuracy of \texttt{AET-SGD} with different number of iterations $K$ and $5$ compute nodes}
\label{tbl:aet_num_iteration}

\end{table}

Our observation from Table~\ref{tbl:aet_num_iteration} is that when the number of iterations is {about from $50,000$ to $75,000$}, we gain the highest accuracy. In addition, if we continue to increase {the number of iterations to $K=100,000$ for example}, the accuracy keeps nearly unchanged, i.e, a larger number of iterations does not improve accuracy any further.
This is because the test accuracy is measured with respect to a certain (test) data set of samples from distribution ${\cal D}$ for which the fraction of correct output labels is computed. Such a fixed test data set introduces an approximation error with respect to the training data set; the training accuracy which is minimized by minimizing the objective function is different from the test accuracy. 
Therefore, it does not help to attempt to converge closer to the global minimum than the size of the approximation error. Hence, going beyond a certain number of iterations will not reduce the estimated objective function (by using the test data set) any further.

\subsection{\texttt{AET-SGD} with different \# of compute nodes}
\label{sec:compute_node}
We want to understand how varying the number  $n$ of compute nodes while fixing other parameters, such as the total number of iterations $K_{total}=60,000$ for all compute nodes (i.e, each compute node will have around $K = \ceil{\frac{K_{total}}{n}}$ SGD iterations)  and diminishing step size sequence $\eta_t$, affects the test accuracy. 
The goal of this experiment is to show that the number of clients $n$ can not be arbitrary large due to the restriction from the delay function $\tau$.

To make the analysis simple, we consider 
asynchronous SGD 
with $d=1$ (i.e., each compute node is allowed to run faster than its neighbors for at most $1$ communication round) and unbiased data sets. Moreover, we choose a linearly increasing sample size sequence $s_{i,c}=a \cdot i^p + b$, where $p=1$ and $a=10, b=0$ for non-convex problems. The experiment uses an initial step size $\eta_{0}=0.01$ with diminishing round step size sequence corresponding to $\eta_t = \frac{\eta_0}{1 + \beta \cdot \sqrt{t}}$ for non-convex problems. Note that we use the iid data sets for each compute node in this experiment.

\begin{table}[ht!]
\begin{center}
\scalebox{0.9}{
\begin{threeparttable}
\begin{tabular}{|c|c|c|c|c|}
\hline
Data set & \# of node & Accuracy & Duration & Speedup \\ 
\hline
\hline
\multirow{5}{*}{MNIST} & $1$ $^{\dag}$  &   $0.9883$   &     $182 ^{\ddag}$  & $1.0^{*}$
\\ 
\cline{2-5}
& $2$   &   $0.9906$       &    $132$      & $1.38$
\\ 
\cline{2-5}
& $5$   &  $0.9881$      &     $121$   & $1.50$
\\ 
\cline{2-5}
& $10$        &   $0.9850$       &     $110$ & $1.65$
\\ 
\cline{2-5}
& $15$        &  $0.9837$       &     $109$   & $1.67$
\\ 
\hline
\hline

\multirow{5}{*}{FashionMNIST} & $1$  &   $0.8947$   &     $182$ & $1.0$
\\ 
\cline{2-5}
& $2$   &   $0.8932$       &    $129$     & $1.41$
\\ 
\cline{2-5}
& $5$   &  $0.8795$      &     $122$   & $1.49$
\\ 
\cline{2-5}
& $10$        &   $0.8737$       &     $114$  & $1.60$
\\ 
\cline{2-5}
& $15$        &  $0.8794$       &     $113$    & $1.61$
\\ 
\hline
\hline

\multirow{5}{*}{KMNIST} & $1$  &   $0.9432$   &     $184$ & $1.0$
\\ 
\cline{2-5}
& $2$   &   $0.9435$       &    $130$     & $1.42$
\\ 
\cline{2-5}
& $5$   &  $0.9352$      &     $126$   & $1.46$
\\ 
\cline{2-5}
& $10$        &   $0.9350$       &     $112$ & $1.64$
\\ 
\cline{2-5}
& $15$        &  $0.9336$       &     $115$     & $1.60$
\\ 
\hline
\hline

\multirow{5}{*}{CIFAR10} & $1$ & $0.6657$ & $3061$  &   $1.0$
\\ 
\cline{2-5}
& $2$   &   $0.6634$       &    $2414$     & $1.27$
\\ 
\cline{2-5}
& $5$   &  $0.6446$      &     $2172$   & $1.41$
\\ 
\cline{2-5}
& $10$        &   $0.6433$       &     $2053$ & $1.50$
\\ 
\cline{2-5}
& $15$        &  $0.6256$       &     $1859$   &  $1.65$
\\ 
\hline

\end{tabular}
 \begin{tablenotes}
        \item {\footnotesize $\dag$ When the number of compute node $n=1$, \texttt{AET-SGD} works as the traditional SGD method.}
 
      \item {\footnotesize $\ddag$ Duration means the whole training time in second unit when the compute nodes start their computation process until they finish their work, including the data processing time in each compute node.}
      
      \item {\footnotesize $*$ Speedup means the ratio between the duration of $n$ compute nodes and the duration of a single compute node.}
  \end{tablenotes}
  \end{threeparttable}
}
\end{center}

\caption{Test accuracy of \texttt{AET-SGD} with different number of compute nodes $n$ and the total number of iteration $K_{total}=60,000$ for all compute nodes, i.e., $K= \ceil{ \frac{K_{total}}{n} }$ iterations for each compute node, asynchronous round $d=1$}
\label{tbl:aet_diff_node}

\end{table}

{As can be seen from Table~\ref{tbl:aet_diff_node}, when we increase the number of compute nodes $n$, the training duration decreases gradually. Specifically, when $n=1$ (AET-SGD yields SGD with single compute node) we mostly achieve the best accuracy, compared to other settings. When $n=2$ or $n=5$, we get nearly same accuracy, compared to a single SGD setting while reducing the training time significantly. However, if we continue to increase  $n$ to a large number, for example $n=15$, then the accuracy has the trend to decrease and training duration starts to reach a limit. In other words, the training duration seems not to decrease with respect to the number of compute nodes. Next, we note that we are more likely to have a slow compute node among $n$ nodes if $n$ is large. The slowest compute node out of $n$ nodes is expected to be slower for increasing $n$ because it needs to transmit/receive more gradient updates to/from others. This means that other compute nodes will need to start waiting for this slowest compute node (see the while $\textsc{Synchronize}({H}, d, i) == \textsc{True}$ loop in Algorithm~\ref{alg:client_adaptive}, which waits for the compute nodes to transmit or receive a broadcast message to or from the the slower neighbors). So, a reduction in execution time due to parallelism among a larger number $n$ of compute nodes will have less of an effect. For increasing $n$, the execution time (duration) will reach a lower limit (i.e., there is a speedup saturation in the distributed learning framework). Another observation is that the speedup ratio is about $1.6x$, compared to the single compute node. This can explained by a strict asynchronous round $d=1$, which forces the compute nodes to wait for each other and eventually increase the training time in total. To answer this question clearly, we conduct another experiment with different values of $d$ in the next experiment.

}

\subsection{\texttt{AET-SGD} with different number of asynchronous communication rounds}
\label{sec:asynchronous_round}

The purpose of this set of experiments is try to investigate the effectiveness of the asynchronous round $d$ to the final test accuracy. We will vary the values of $d$ from $1$ to $10$ while keeping the number of iteration $K=60,000$ for each compute node. To make the analysis simple, we choose a linearly increasing sample size sequence $s_{i,c}=a \cdot i^p + b$, where $p=1$ and $a=10, b=0$ for non-convex problems. The experiment uses an initial step size $\eta_{0}=0.01$ with diminishing round step size sequence corresponding to $\eta_t = \frac{\eta_0}{1 + \beta \cdot \sqrt{t}}$. Other hyper parameters such as $\eta_0, \beta, n$ and the network topology are the same as in
Table~\ref{tbl:basic_param_setting}.

\begin{table}[ht!]
\begin{center}
\scalebox{0.9}{
\begin{tabular}{|c|c|c|c|}
\hline
Data set & Async. round $d$  & Accuracy & Duration \\ \hline
\hline
\multirow{6}{*}{MNIST}   & $0$  &  $0.9886$        &      $179$    
\\ \cline{2-4} 
                         & $1$  &  $0.9881$        &      $121$
\\ \cline{2-4}
                         & $2$  &  $0.9888$        &    $112$      
\\ \cline{2-4}
                         & $5$  &  $0.9812$        &    $78$      
\\ \cline{2-4}
                         & 7  &    $0.9768$      &      $62$    
\\ \cline{2-4}
                         & 10 &    $0.9612$      &      $60$    
\\ \hline
\hline

\multirow{6}{*}{FashionMNIST}   & $0$  &  $0.8886$        &      $189$    
\\ \cline{2-4} 
                         & $1$  &  $0.8868$        &      $122$
\\ \cline{2-4}
                         & $2$  &  $0.8823$        &    $105$      
\\ \cline{2-4}
                         & $5$  &  $0.8814$        &    $82$      
\\ \cline{2-4}
                         & 7  &    $0.8768$      &      $65$    
\\ \cline{2-4}
                         & 10 &    $0.8212$      &      $64$    
\\ \hline
\hline

\multirow{6}{*}{KMNIST}   & $0$  &  $0.9412$        &      $180$  
\\ \cline{2-4} 
                         & $1$  &  $0.9352$        &      $126$
\\ \cline{2-4}
                         & $2$  &  $0.9323$        &    $115$      
\\ \cline{2-4}
                         & $5$  &  $0.9244$        &    $78$      
\\ \cline{2-4}
                         & 7  &    $0.9216$      &      $70$    
\\ \cline{2-4}
                         & 10 &    $0.9221$      &      $67$    
\\ \hline
\hline

\multirow{5}{*}{CIFAR10} & $0$  &    $0.6486$      &  $2792$
\\ \cline{2-4} 
                        & $1$  &    $0.6446$      &  $2172$  
\\ \cline{2-4} 
                         & $2$  &   $0.6432$       &    $1786$      
\\ \cline{2-4} 
                         & $5$  &   $0.6398$       &    $1265$      
\\ \cline{2-4} 
                         & $7$  &   $0.6331$       &    $890$      
\\ \cline{2-4} 
                         & $10$ &   $0.6019$       &    $868$      
\\ \hline
\end{tabular}
}
\end{center}

\caption{Test accuracy of \texttt{AET-SGD} with different values of asynchronous round $d$}
\label{tbl:asymchrnous_round}

\end{table}

As can be seen from Table~\ref{tbl:asymchrnous_round}, when we increase the number of asynchronous round $d$, the training duration (in second) will decrease dramatically. For example, when the value of asynchronous $d=7$, we get about $2.74x$ and $3.14x$ speedup (comparing with the baseline $d=0$ when all the compute nodes need to wait for each other) for MNIST and CIFAR$10$ dataset respectively. We extend the experiments for FashionMNIST and KMNIST data sets and observe the same pattern. Note that the test accuracy goes down when we increase the number of asynchronous $d$. Specifically, when $d=1$ to $d=7$, the loss of test accuracy is negligible, compared to the test accuracy when $d=10$. This answers the question how can we speed up the training process or enlarge the level of speedup saturation in the decentralized learning architecture. We need to carefully select the number of compute node $n$ and the number of asynchronous round $d$ in practice in order to obtain a desired speedup ratio.

{\em Our experimental results suggest that we could choose the asynchronous round $d$ upto $7$ communication rounds and deploy the training system with about $10$ compute nodes or more for securing a good test accuracy and desired level of speedup.}

\end{document}